\documentclass{article}

\usepackage{arxiv}

\usepackage[utf8]{inputenc} 
\usepackage[T1]{fontenc}    
\usepackage{hyperref}       
\usepackage{url}            
\usepackage{booktabs}       
\usepackage{amsfonts}       
\usepackage{nicefrac}       
\usepackage{microtype}      
\usepackage{lipsum}
\usepackage[colorinlistoftodos,prependcaption,textsize=tiny]{todonotes}
\usepackage{xargs} 
\usepackage[numbers]{natbib}

\usepackage{soul}
\usepackage{subfig}
\usepackage{graphicx}
\usepackage{multirow}
\usepackage{csquotes}
\usepackage{multirow}
\usepackage{rotating} 
\usepackage{xcolor}
\usepackage{amssymb}
\usepackage{amsmath}
\usepackage{array}
\usepackage[font=scriptsize]{caption}

\usepackage[font=small]{caption}
\usepackage{csquotes}
\usepackage{tabularx}

\title{Scraping Social Media Photos Posted in Kenya and Elsewhere to Detect and Analyze Food Types}

\author{
  Kaihong Wang \thanks{The first and second authors contributed equally to this research.}\\
  Department of Computer Science\\
  Boston University\\
  Boston, MA 02215 \\
  \texttt{kaiwkh@bu.edu} \\
  \And
  Mona Jalal \footnotemark[1]\\
  Department of Computer Science\\
  Boston University\\
  Boston, MA 02215 \\
  \texttt{jalal@bu.edu} \\
  \And
  Sankara Jefferson \\
  Lori Systems LTD\\
  Nairobi, Kenya\\
  \texttt{jefferson.sankara@lorisystems.com} \\
  \And
  Yi Zheng \\
  Department of Computer Science\\
  Boston University\\
  Boston, MA 02215 \\
  \texttt{yizheng@bu.edu} \\
  \And
  Elaine O. Nsoesie \\
  Department of Global Health\\
  Boston University\\
  Boston, MA 02215 \\
  \texttt{onelaine@bu.edu} \\
  \And
  Margrit Betke \\
  Department of Computer Science\\
  Boston University\\
  Boston, MA 02215 \\
  \texttt{betke@bu.edu} \\
  }

\begin{document}
\maketitle

\begin{abstract}
Monitoring population-level changes in diet could be useful for education and for implementing interventions to improve health. Research has shown that data from social media sources can be used for monitoring dietary behavior. We propose a {\bf scrape-by-location} methodology to create food image datasets from Instagram posts.
We used it to collect 3.56 million images over a period of 20 days in March 2019.  We also propose a {\bf scrape-by-keywords} methodology and used it to scrape $\sim$30,000 images and their captions of 38 Kenyan food types. We publish two datasets of 104,000 and 8,174 image/caption pairs, respectively. With the first dataset, {\bf Kenya104K},  we train a {\bf Kenyan F}ood {\bf C}lassifier, called {\bf KenyanFC}, to distinguish Kenyan food from non-food images posted in Kenya.  We used the second dataset, {\bf KenyanFood13}, to train a classifier {\bf KenyanFTR}, short for {\bf Kenyan F}ood {\bf T}ype {\bf R}ecognizer, to recognize 13 popular food types in Kenya. 
The KenyanFTR is a multimodal deep neural network that can identify 13 types of Kenyan foods using both images and their corresponding captions. Experiments show that the average top-1 accuracy of KenyanFC is 99\% over
10,400 tested Instagram images and of KenyanFTR is 81\% over 8,174 tested data points.
Ablation studies show that three of the 13 food types are particularly difficult to categorize based on image content only and that adding analysis of captions to the image analysis yields a classifier that is 9 percent points more accurate than a classifier that relies only on images.  Our food trend analysis revealed that cakes and roasted meats were the most popular foods in photographs on Instagram in Kenya in March 2019. 
\end{abstract}


\section{Introduction}

Data from Instagram can be used to study a community's food consumption patterns as people post images of what and where they eat. Instagram users share their activities, moods, and location of particular venues in real time, thereby allowing for aspects of their experiences and patterns of eating and drinking to be captured. This enables the exploration of questions such as, why do people share images on social media and what types of foods do people tend to share on social media~\cite{DechoudhuryShKi16,PhanGa17}. Instagram hashtags have been shown to be useful in the study of the characteristics of food as well as the context and interests of users related to food~\cite{MejovaAbHa16}. Images and hashtags together can provide insights into how particularly youths perceive and interact with food in their communities, allowing researchers to study the diverse aspects of food culture in a specific demographical group. Furthermore, data on food and location from Instagram can allow researchers to establish eating patterns and their possible associations to conditions like obesity~\cite{BlanchardMa2007retail,PriceCrAmKaMuTaBrLaMwNk2018}. 

\begin{figure}[ht!]
\centering
\captionsetup[subfigure]{labelformat=empty}
   \subfloat[ \label{fig:PKR}]{%
      \includegraphics[width=4cm, height=4cm]{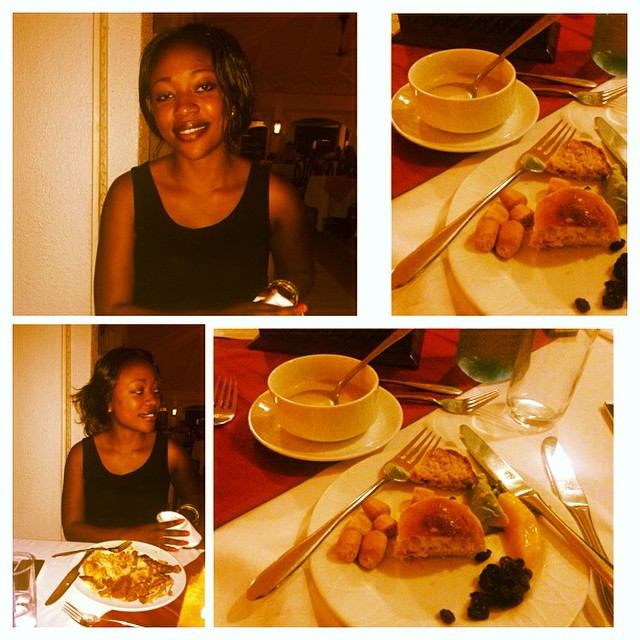}}
\hspace{2.5em}
   \subfloat[ \label{fig:PKT} ]{%
      \includegraphics[width=4cm, height=4cm]{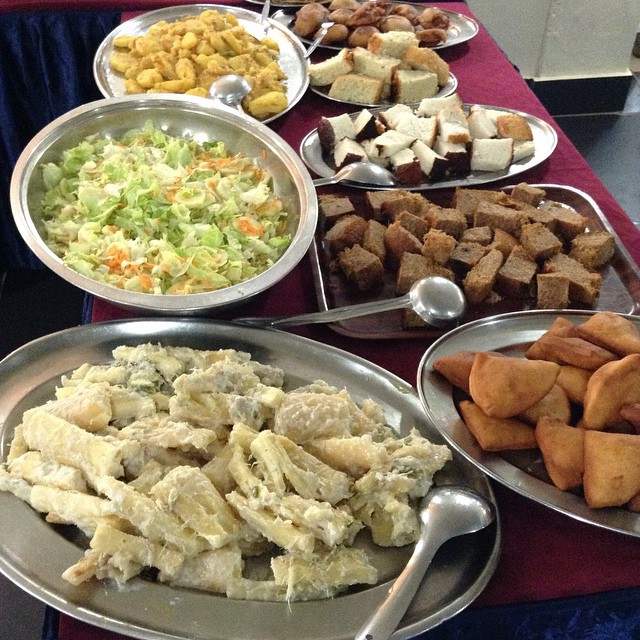}}
\hspace{2.5em}
   \subfloat[ \label{fig:tie5}]{%
      \includegraphics[width=4cm, height=4cm]{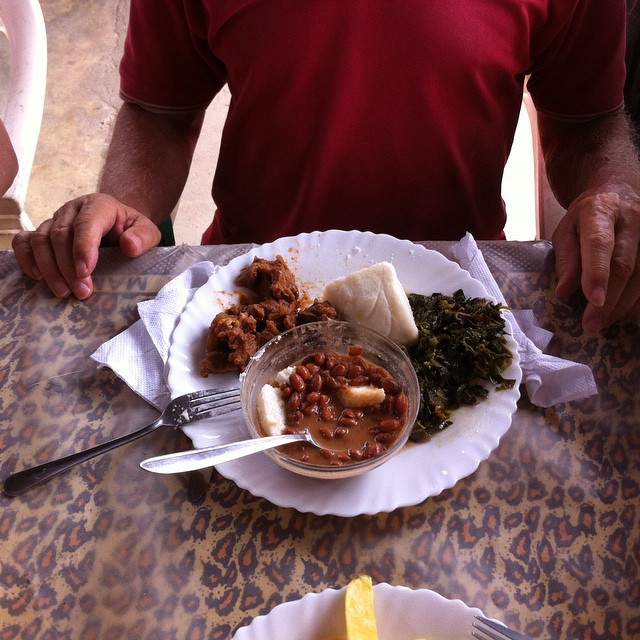}}
\\[-1ex]
      \subfloat[ \label{fig:PKR}]{%
      \includegraphics[width=4cm, height=4cm]{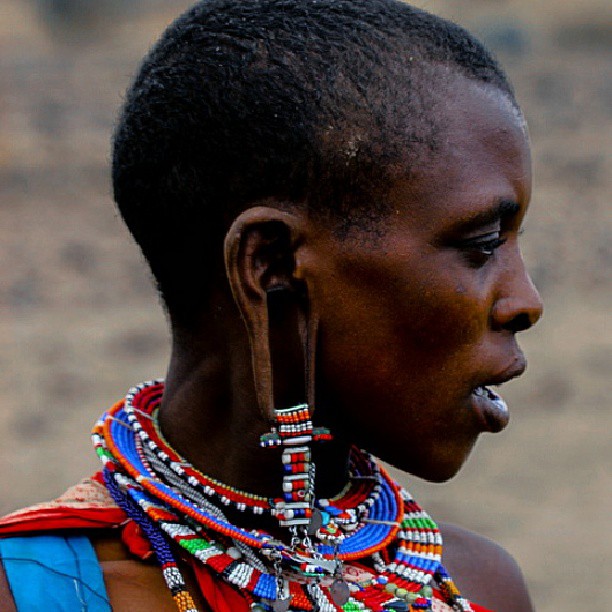}}
\hspace{2.5em}
   \subfloat[ \label{fig:PKT} ]{%
      \includegraphics[width=4cm, height=4cm]{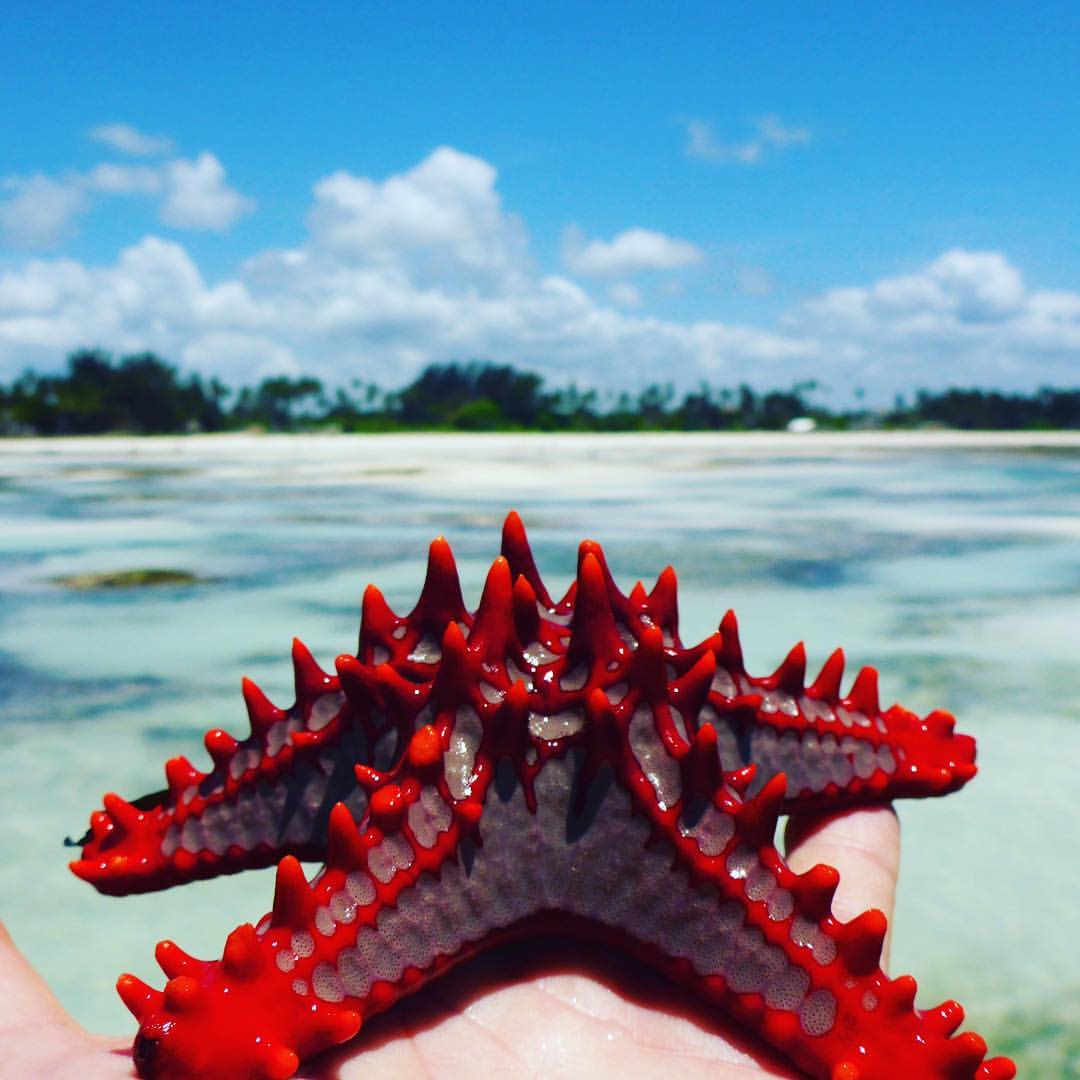}}
\hspace{2.5em}
   \subfloat[ \label{fig:tie5}]{%
      \includegraphics[width=4cm, height=4cm]{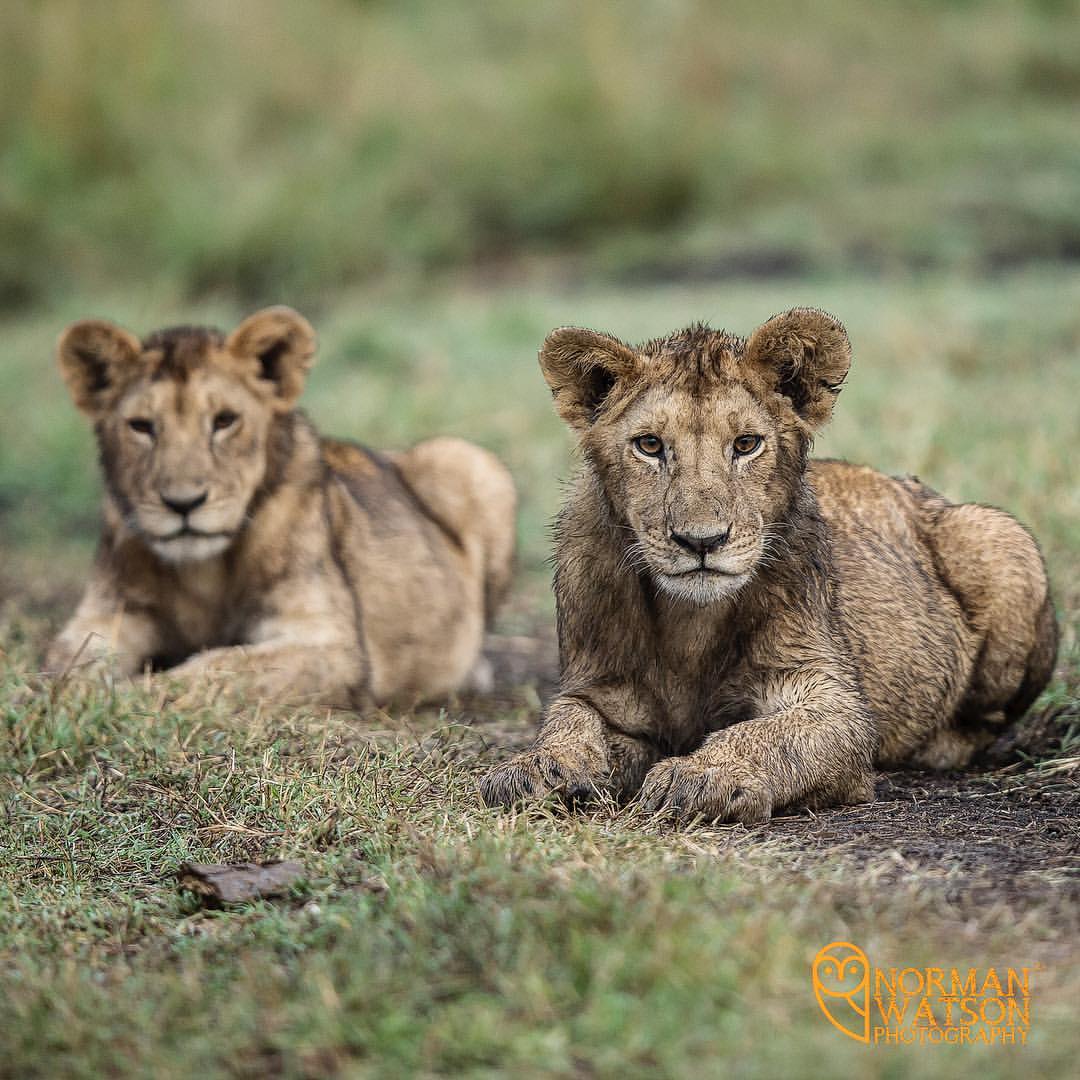}}
      \\[-2ex]

\caption{\small Sample images of the proposed Kenya104K dataset.}
    \label{fig:sample_images}
\end{figure}

Our work is unique in its focus on an African country (Fig.~\ref{fig:sample_images}). There is a dearth of studies evaluating the use of social media for studying nutrition and diet in African countries using computer vision algorithms. We take advantage of the wide adoption of mobile technology and social media in African countries, especially in urban areas to provide tools that enable social scientists to assess the utility of social media data for studying diets and attitudes towards foods in Kenya. Specifically, in this paper, we describe methods to produce large-scale datasets that can be used to study food trends on Instagram.  We then apply these methods to collect and analyze datasets of Kenyan food images and their captions.  

\begin{figure*}[h!]
\centering
\captionsetup[subfigure]{labelformat=empty}
   \subfloat[ Bhaji, 789\label{fig:PKR}]{%
      \includegraphics[width=3cm, height=3cm]{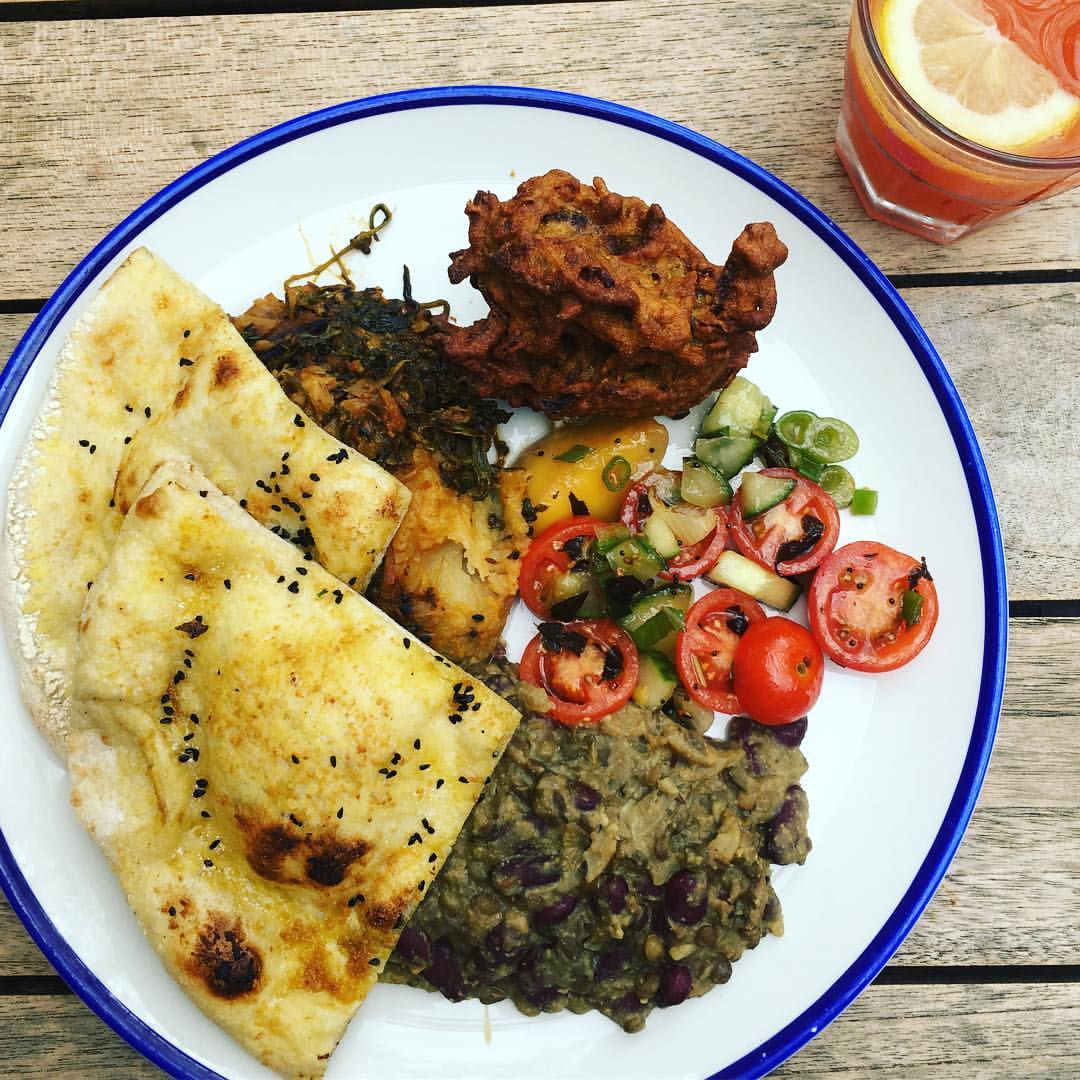}}
\hspace{.5em}
   \subfloat[Chapati, 1,076 \label{fig:PKT} ]{%
      \includegraphics[width=3cm, height=3cm]{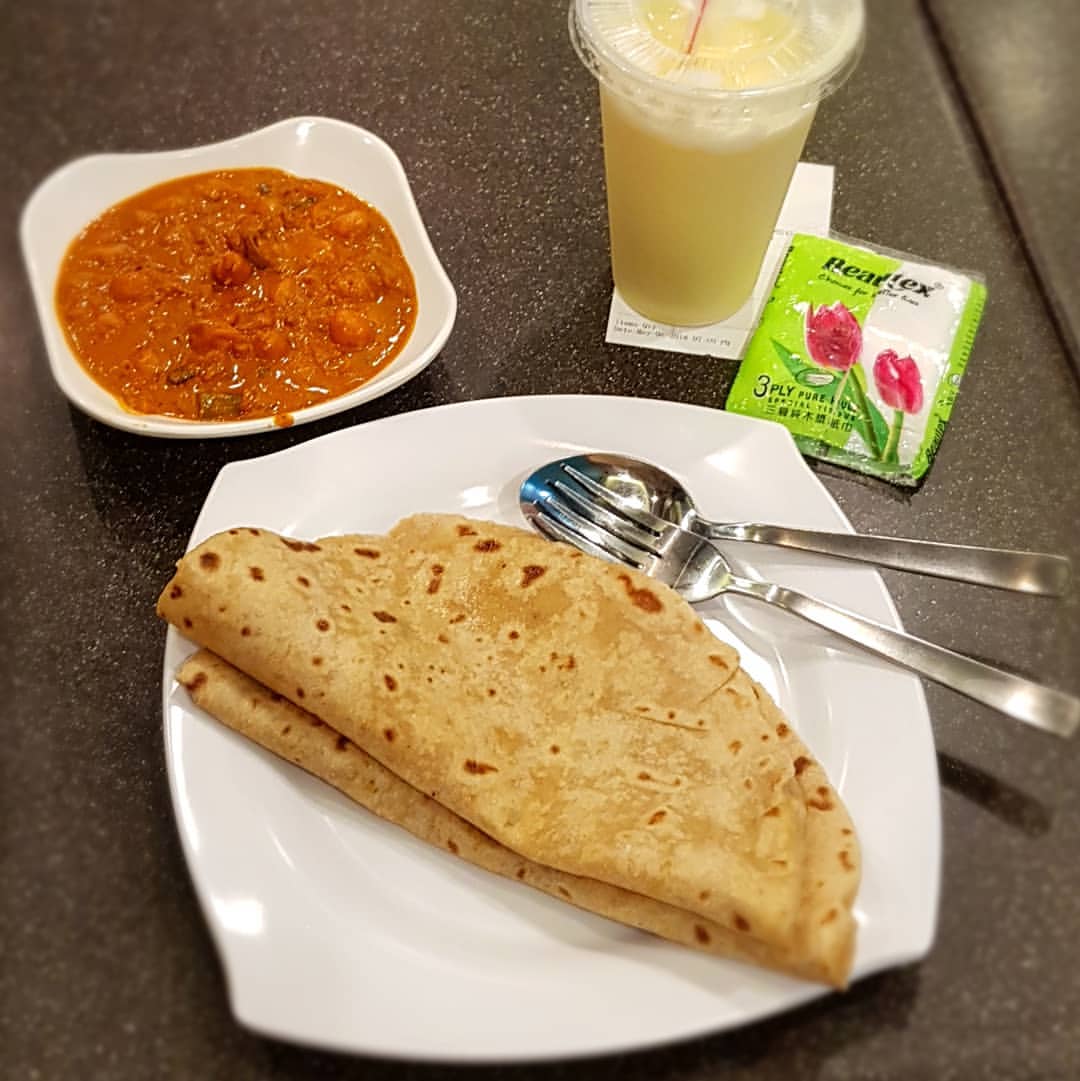}}
\hspace{.5em}
   \subfloat[Nyama choma, 980 \label{fig:tie5}]{%
      \includegraphics[width=3cm, height=3cm]{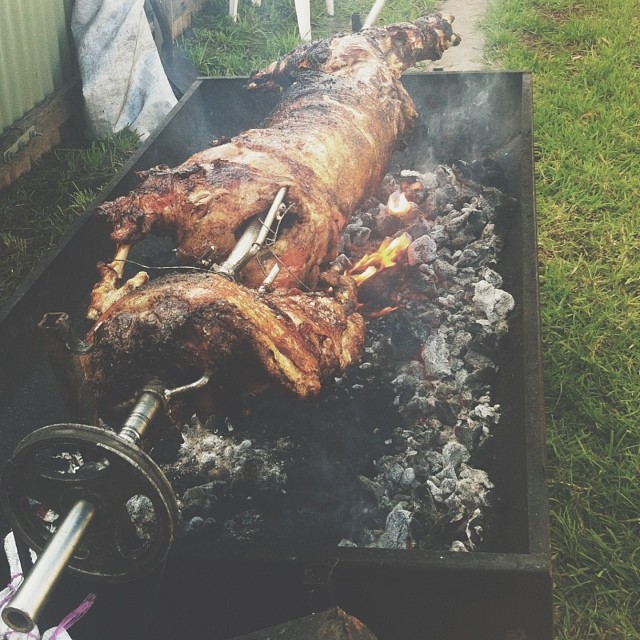}}
\hspace{.5em} 
    \subfloat[Mandazi, 775\label{fig:PKR}]{%
      \includegraphics[width=3cm, height=3cm]{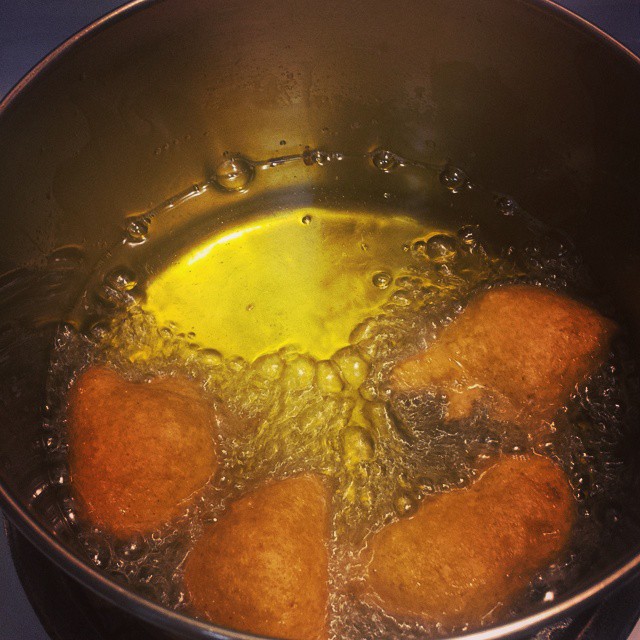}}\\

      \subfloat[Masala chips, 546 \label{fig:PKR}]{%
      \includegraphics[width=3cm, height=3cm]{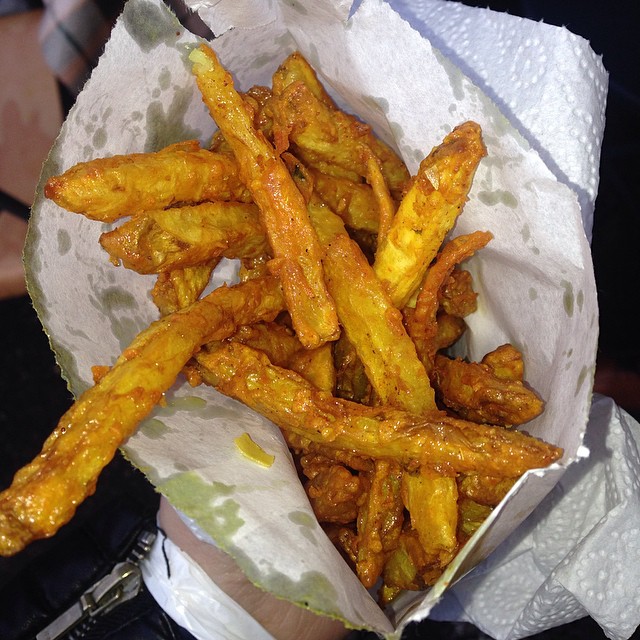}}
\hspace{.5em}
   \subfloat[Kachumbari, 619 \label{fig:PKT} ]{%
      \includegraphics[width=3cm, height=3cm]{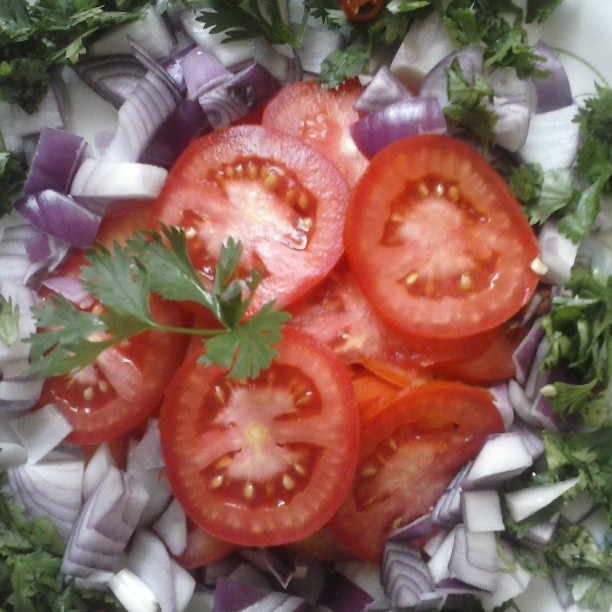}}
\hspace{.5em}
   \subfloat[Ugali, 785 \label{fig:tie5}]{%
      \includegraphics[width=3cm, height=3cm]{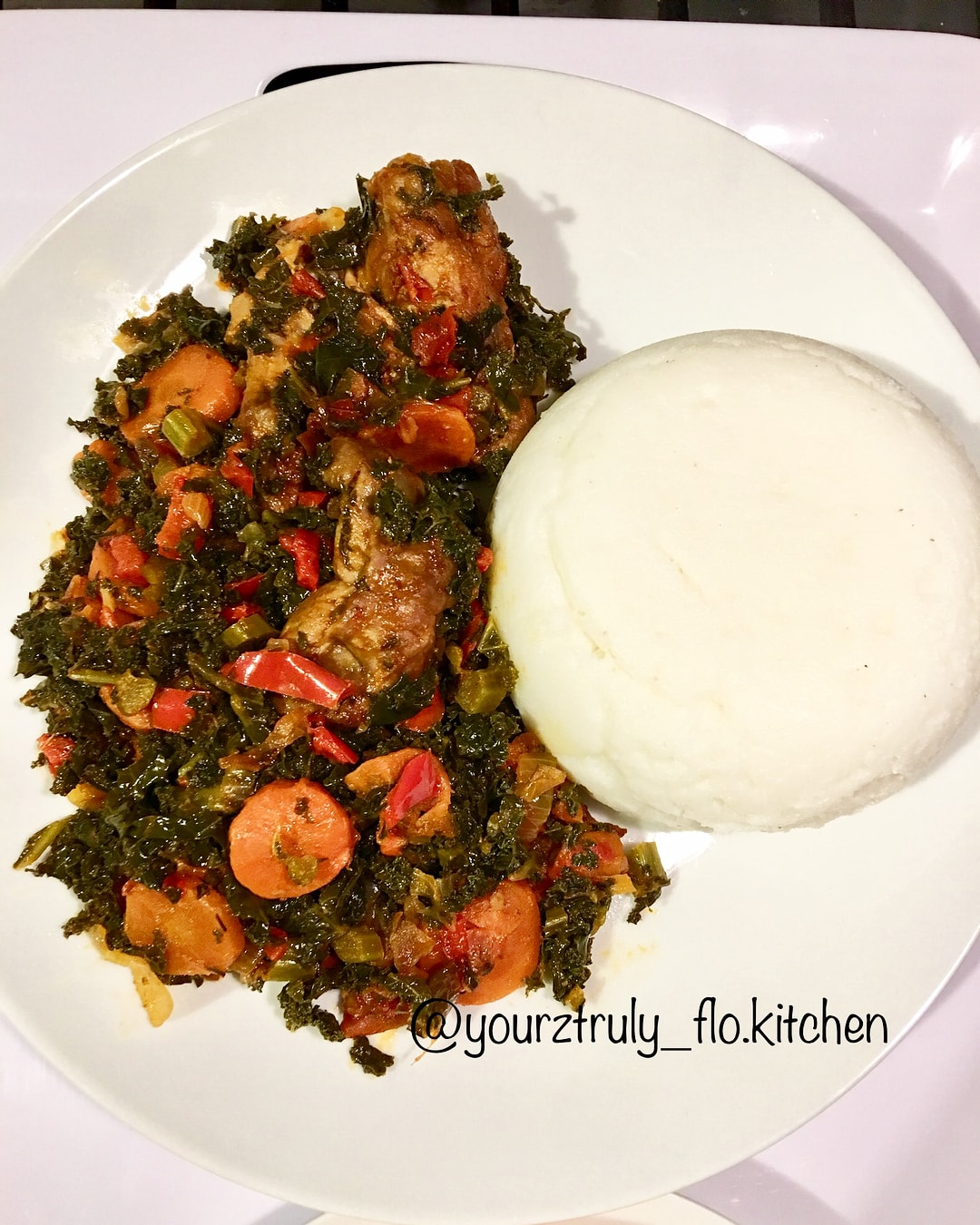}}
\hspace{.5em}
    \subfloat[Pilau, 410 \label{fig:PKR}]{%
      \includegraphics[width=3cm, height=3cm]{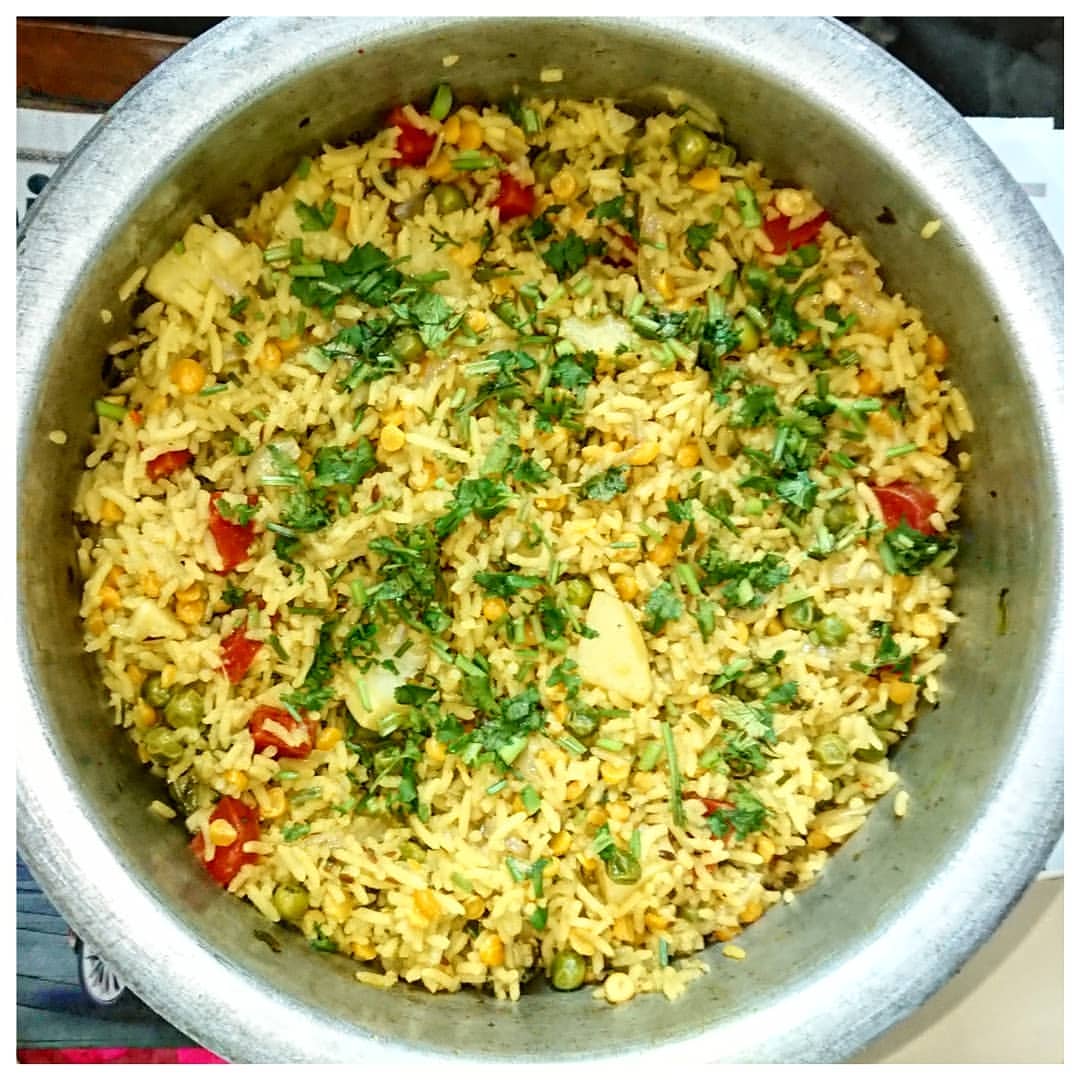}}\\


      \subfloat[Matoke, 604 \label{fig:PKR}]{%
      \includegraphics[width=3cm, height=3cm]{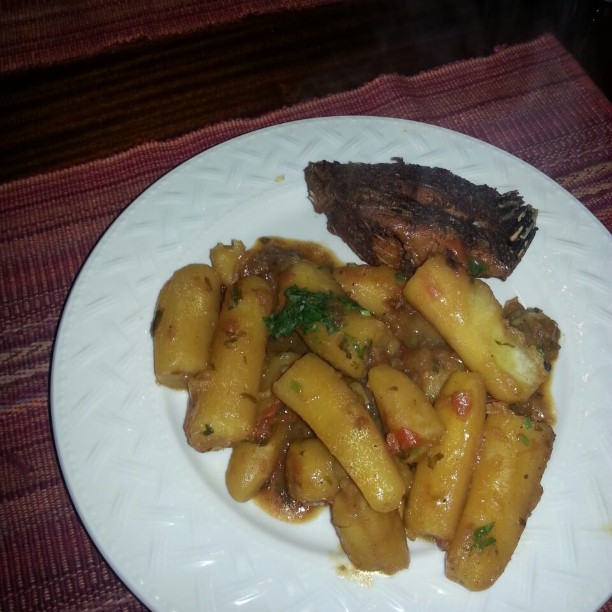}}
\hspace{.5em}
   \subfloat[Githeri, 600 \label{fig:PKT} ]{%
      \includegraphics[width=3cm, height=3cm]{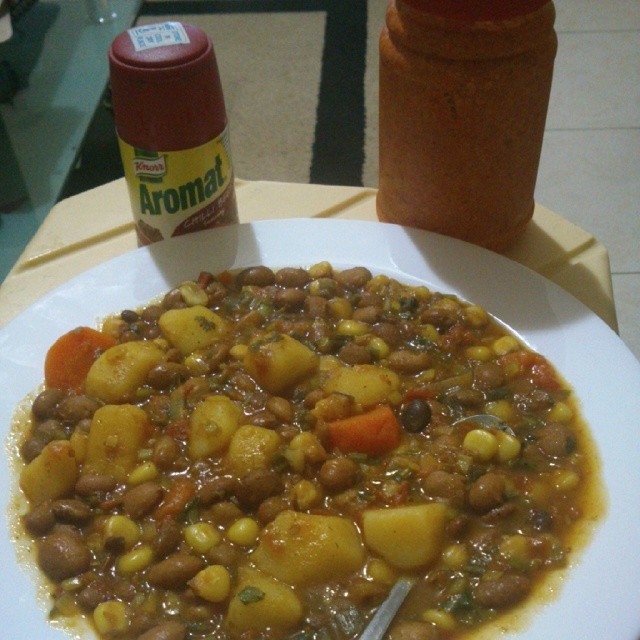}}
\hspace{.5em}
   \subfloat[Mukimo, 266 \label{fig:tie5}]{%
      \includegraphics[width=3cm, height=3cm]{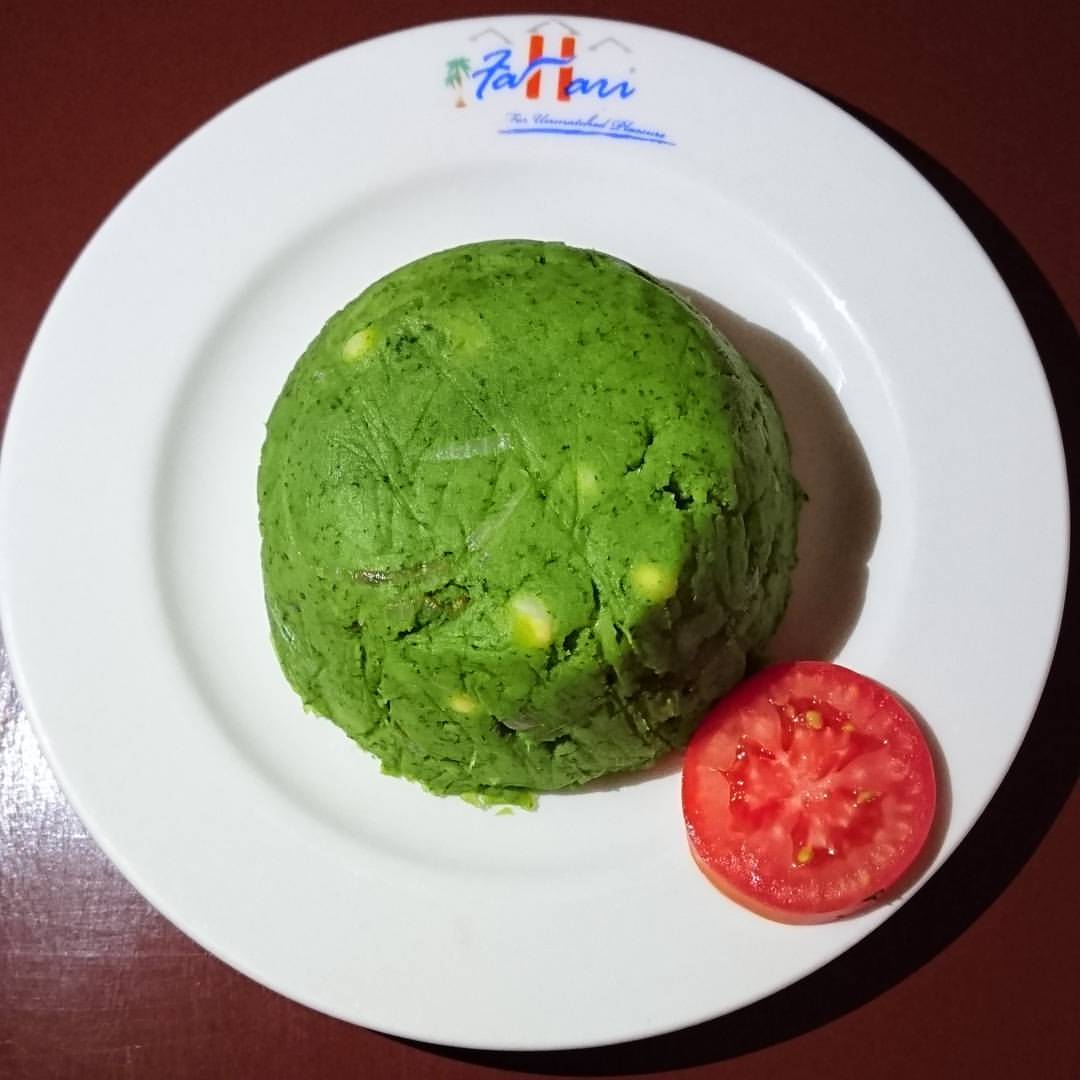}}
\hspace{.5em}
    \subfloat[Sukuma wiki, 505\label{fig:PKR}]{%
      \includegraphics[width=3cm, height=3cm]{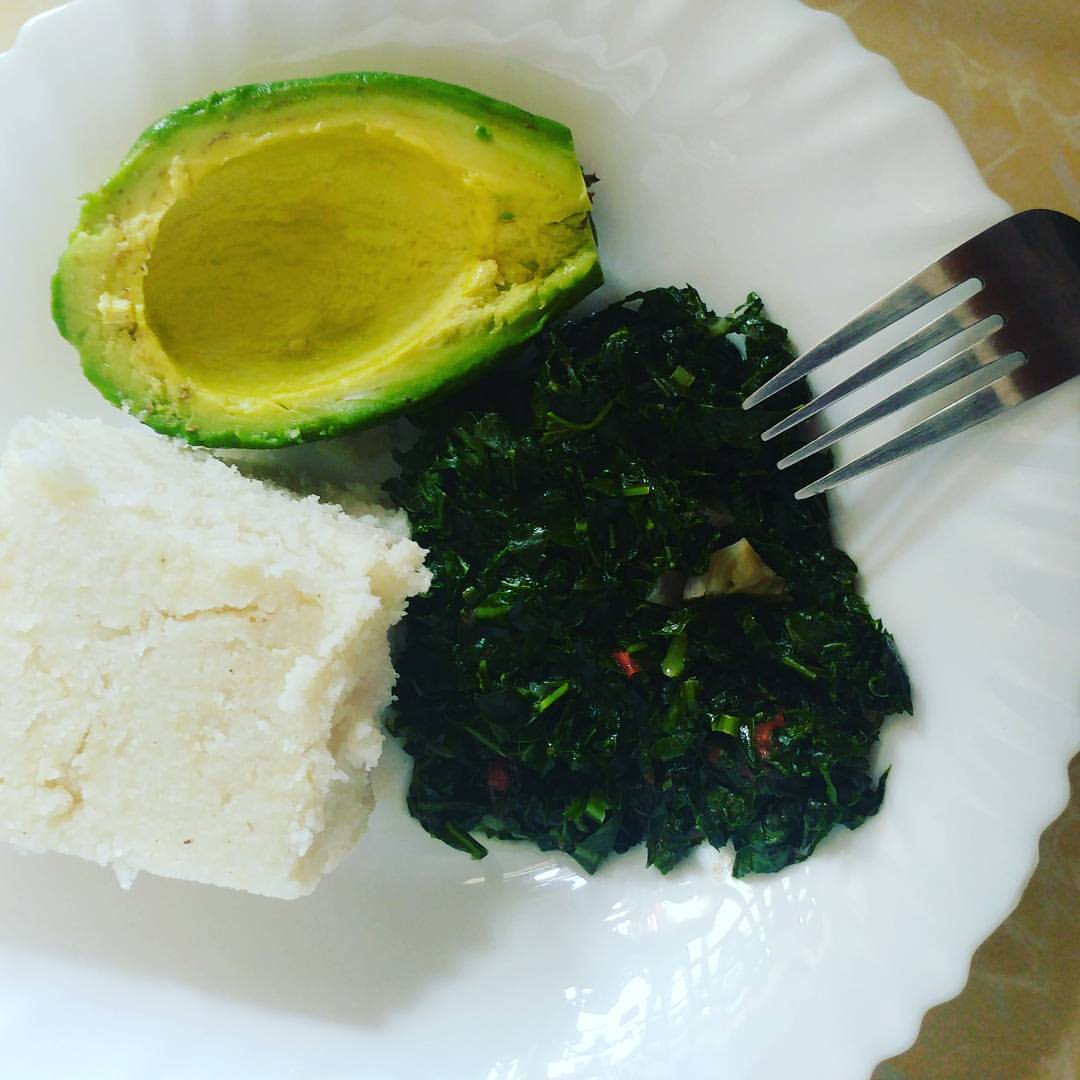}} \hspace{.5em}
          \subfloat[Kuku choma, 219\label{fig:PKR}]{%
      \includegraphics[width=3cm, height=3cm]{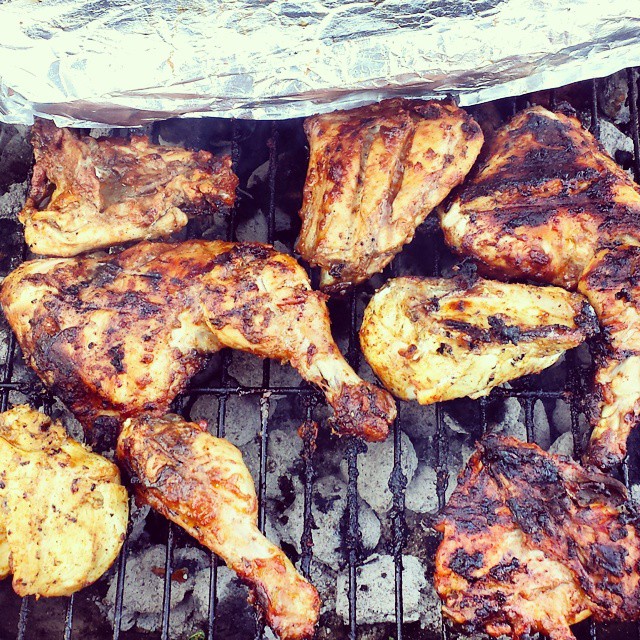}}
\caption{\small Sample images  of Kenyan food types in the proposed KenyanFood13 dataset with numbers of samples for each type.   }
    \label{fig:type_example}
\end{figure*}

Kenya has a dietary culture that is distinct from Western countries, so we could not simply apply previously developed food type detectors on Kenyan Instagram posts. In order to develop our own Kenyan food type detector, we had to overcome the challenge of automating the scraping and collecting of training images 
with labels that define the food type.  Another challenge was to effectively scrape and filter extremely large numbers of Instagram images from a specific geographic region.  
Since we downloaded all Kenya-originated images from Instagram during a twenty day period in March 2019, we can use our classifier (as well as existing classifiers for foods not specific to Kenya) to analyze the prevalence of food images as well as food types (Fig.~\ref{fig:type_example}) in Kenyan Instagram posts during that period.

As far as we know, there is no publicly available food image dataset dedicated to the task of Kenyan food recognition. 
Our efforts to create Kenyan-specific datasets and classifiers are crucial for any food trend analysis a health informaticist may want to conduct for Kenya.
Dibia et al.~\cite{dibiacoco} and Buolamwini et al.~\cite{BuolamwiniGe18} warned that the generalization of learning systems is undermined when they are trained on datasets without African content or context in African scenarios. 

Our results show, while food-detection classifiers trained on food image datasets that only consist of images of Western or East Asian foods can recognize whether food is shown in an image relatively accurately, results improve when a classifier is trained specifically on Kenyan images. For the task of 
food type recognition, any supervised learning system needs to have examples of specific Kenyan foods to be able to recognize them (and food like ``sukuma wiki'' is not represented in previous datasets).

\noindent
To summarize, our contributions are as follows:


\begin{itemize}
    \item We propose two systems for collecting Instagram posts, a {\bf Scrape-by-\-Location} system and a {\bf Scrape-by-\-Key\-words} system to collect posts about Kenyan foods uploaded in Kenya and elsewhere.
    \item We used our scraping systems to collect millions of Instagram images and their captions, and then devised methods to filter these posts in order to create two datasets.   
    Our multimodal dataset,
    {\bf Kenya104K}, can be applied to train machine learning systems to detect Kenyan foods in images.  Our Kenyan food type dataset, {\bf KenyanFood13}, contains 13 types of popular and representative Kenyan foods.
    \item We propose a {\bf multimodal deep learning model, KenyanFTR,} that interprets feature vectors, which in combination represent food images and their corresponding captions, to predict the type of Kenyan food shown in an image.  We also make a classifier, {\bf KenyanFC}, available that is trained to distinguish food (both Kenyan and non-specific food) from non-food images.
    \item  We applied our techniques to the millions of Instagram posts we collected across Kenya over a period of 20 days to give an example of the kind of analysis social scientists may conduct with our tools.
\end{itemize}

Our datasets and code is publicly available at {\tt https://github.com/monajalal/Kenyan-Food}. We hope that our work encourages other researchers to include images from Africa in their image analysis research by either following our methodology to create their own Instagram datasets or using our Kenyan image data to further develop computer vision tools to recognize and study image content.

\section{Related Work}

Studies have shown that image and text data from social media sites, such as Instagram and Twitter, can be useful for monitoring diet ~\cite{FriedSuKoHiBe2014analyzing,MejovaHaNoWe2015foodporn,SharmaDe2015measuring} 
or identifying 
deserts~\cite{DechoudhuryShKi16}. 
Images of food have been used in apps to measure food calories in real time
(e.g., https://caloriemama.ai\st{/}) and to log food 
~\cite{Lu16,TurnerLe2017instagram}.
Various datasets have been collected from social media (e.g., ~\cite{ChenNg16,MinBaMeZhRuJi18,OfliAyWeHaTo17,RichHaHo16,WangKuThCoPr15}).


\subsection{Food/Non-food Datasets and Classification}

Datasets built to support food detection tasks have two classes, namely ``food'' and ``non-food.'' Images in the food class include various kinds of foods while images in the non-food class should cover as many other objects as possible (human portraits, landscape scenes, and other objects) that could appear in use-case datasets in which food is supposed to be detected. Table~\ref{tab:fnf} lists some popular food/non-food datasets as well as our proposed dataset Kenya104K.

\begin{table}[h]
\centering
\caption{\label{tab:fnf}
Food/Non-food datasets }
\begin{tabular}{lrrr}
\hline
Dataset & \multicolumn{1}{l}{\# food images} & \multicolumn{1}{l}{\# non-food im.\ } & \multicolumn{1}{c}{total} \\\hline
Food-5K~\cite{SinglaYuEb16} & 2,500 & 2,500 & 5,000 \\
IFD~\cite{KagayaAi15}      & 4,230 & 5,428 & 9,658 \\
FCD~\cite{KagayaAi15}      & 25,250 & 28,322 & 53,572 \\ 
Flickr-Food ~\cite{FarinellaAlStBa15}      & 4,805 & 0 & 4,805 \\ 
Flickr-NonFood ~\cite{FarinellaAlStBa15}   & 0 & 8,005 & 8,005 \\ 
{\bf Kenya104K} & 52,000 & 52,000 & 104,000 \\\hline

\end{tabular}

\end{table}

Many methods and models have been proposed for the task of food/non-food classification. 
Early work in analyzing food images by Kitamura et al.~\cite{KitamuraYaAi09} used hand-crafted features such as color histograms, Discrete Cosine Transform coefficients, and shape representations to train a support vector machine (SVM) classifier. Farinella et al.~\cite{FarinellaAlStBa15} used a one-class SVM on features extracted based on a bag-of-words approach to exclude the influence of non-food images. 
More recent research efforts have used deep networks
for food detection and shown their superior performance over traditional approaches
(e.g., ~\cite{KagayaAiOg14,SinglaYuEb16} applied {\em GoogLeNet}~\cite{SzegedyLiJiSeReAnErVaRa15} and {\em Network in Network} ~\cite{LinChYa13}). 
Ragusa et al.~\cite{RagusaToFuBaFa16} 
investigated how to optimally combine different representation methods with classification schemes.

\subsection{Food Type Datasets and Classification}

The task of recognizing food types in images is typically solved with a supervised learning model and under the assumption that images are known to contain food.   
For training food classification models, researchers have proposed many food type datasets collected in various ways. For example, Bossard et al.~\cite{BossardGuGo14} built the ETHZ Food-101 dataset by collecting food images from http://foodspotting.com and randomly sampling 1,000 images from the top 101 most popular dishes with consistent names ranked on the website. Table~\ref{tab:food-type} lists some popular food/non-food datasets,  as well as our proposed dataset KenyanFood13.
\begin{table}[h]
\centering
\caption{\label{tab:food-type}
Food type datasets}
\begin{tabular}{lrrrl}
\hline
Dataset       & {\centering \# of classes}  & {\centering image per class}  & {\centering Total \# of images}  & {\centering Style of food}
\\ \hline
ETHZ Food-101~\cite{BossardGuGo14} & 101                         & 1,000                            & 101,000       &As, E, Am$^1$     \\
UPMC Food-101~\cite{WangKuThCoPr15} & 101                         & 1,000                            & 101,000      &As, E,  Am      \\
UEC-FOOD-100~\cite{MatsudaHoYa12}  & 100                         & $\sim$90                         & 9,060         &Japanese           \\
UEC-FOOD-256~\cite{KawanoYa14}  & 256                         & $\sim$127                         & 31,397          &Japanese         \\
VireoFood-172~\cite{ChenNg16}  & 172                         & $\sim$641                         & 110,241          &Chinese          \\
UNICT-FD889~\cite{FarinellaAlSt14}   & 889                         & $\sim$4                          & 3,583       &As, E, Am      \\
UNICT-FD1200~\cite{FarinellaAlMoStBa16}   & 1200                         & $\sim$4                          & 4,754       &As, E, Am      \\
Food-524DB~\cite{CioccaNaSc17}    & 524                         & $\sim$473                        & 247,636        &As, E, Am          \\
PFID ~\cite{ChenDhWuYaSuYa09}         & 101                         & 18                               & 1,818      &E, Am              \\
Food500~\cite{MerlerWuUcNgSm16}       & 500                         & $\sim$300                        & 150,000    &As, E, Am              \\ 
NTU-FOOD~\cite{ChenYaHoWaLiChYeOu12}       & 50                         & 100                        & 5,000    &Chinese              \\ 
{\bf KenyanFood13}                                & 13                           & $\sim$629                           & 8,174      &Kenyan  \\\hline
\end{tabular}
\caption*{1 Asian (As),  European (E), American (Am)}

\end{table}

Classic food type recognition models generally follow the pipeline of extracting and combining different features and feeding them into a classifier (e.g., a SVM). For example, Joutou et al.~\cite{JoutouYa09} trained a multiple kernel SVM using a combination of features, including Gabor texture features and color histograms. Bossard et al.~\cite{BossardGuGo14} proposed a method that applies random forests to extract discriminative visual components from the ETHZ Food-101 dataset. Deep neural networks have shown exciting performance in food recognition tasks. Bossard et al.~\cite{BossardGuGo14} showed that an AlexNet~\cite{Krizhevsky14} trained on the ETHZ Food-101 dataset can achieve higher accuracy than other methods they tried. Yanai et al.~\cite{YanaiKa15} showed the effectiveness of fine-tuning a pre-trained AlexNet for food recognition in images. Martinel et al.~\cite{MartinelFoMi16} applied residual learning to the food recognition task by introducing a ``wide-slice residual network.'' %

\section{Methods}

\begin{figure}[]
\centering
\begin{minipage}{.5\textwidth}
  \vspace*{0.4cm}
  \centering
  \includegraphics[width=.9\linewidth]{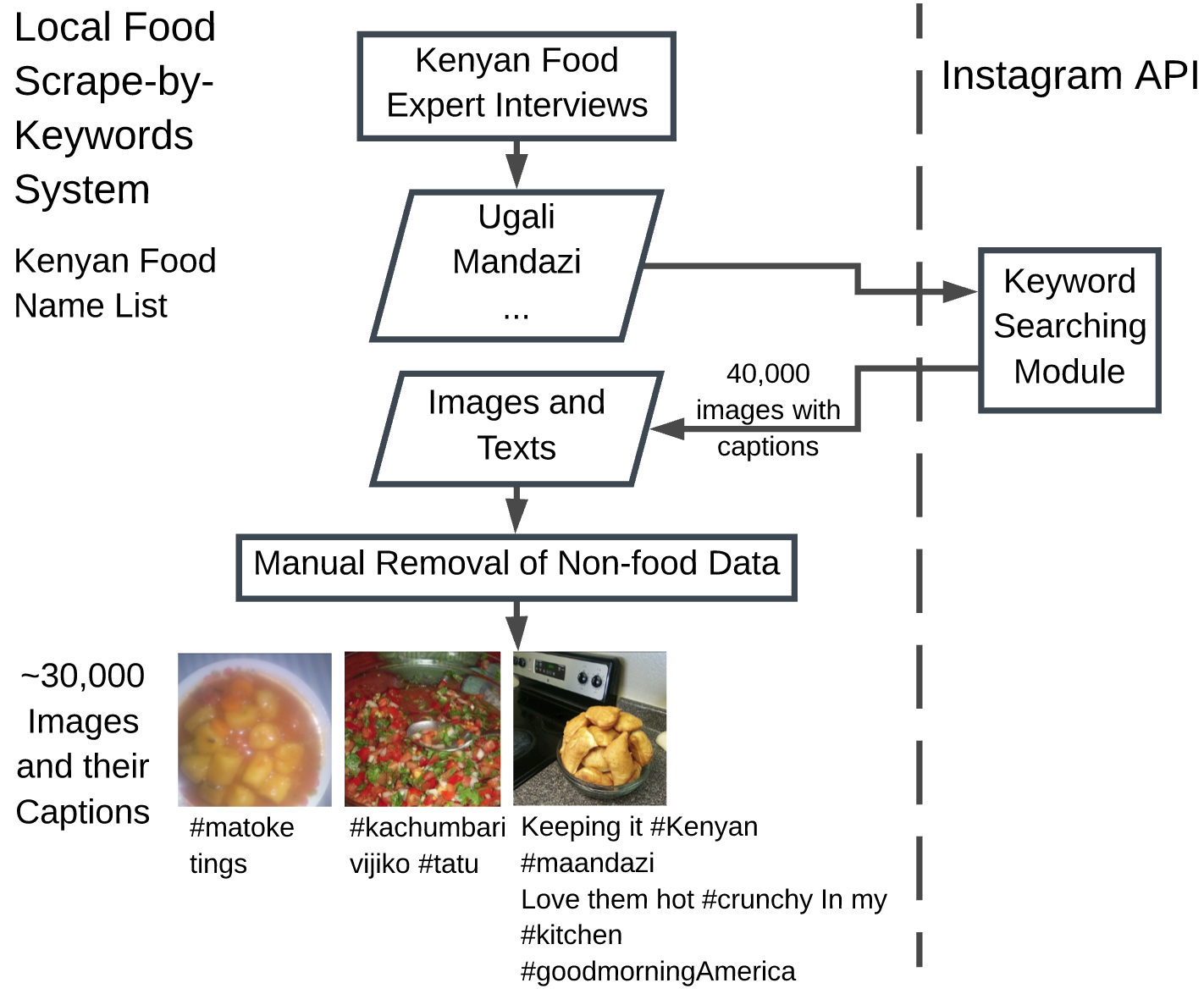}
  \vspace*{1.2cm}
\end{minipage}%
\begin{minipage}{.5\textwidth}
  \centering
  \includegraphics[width=.9\linewidth]{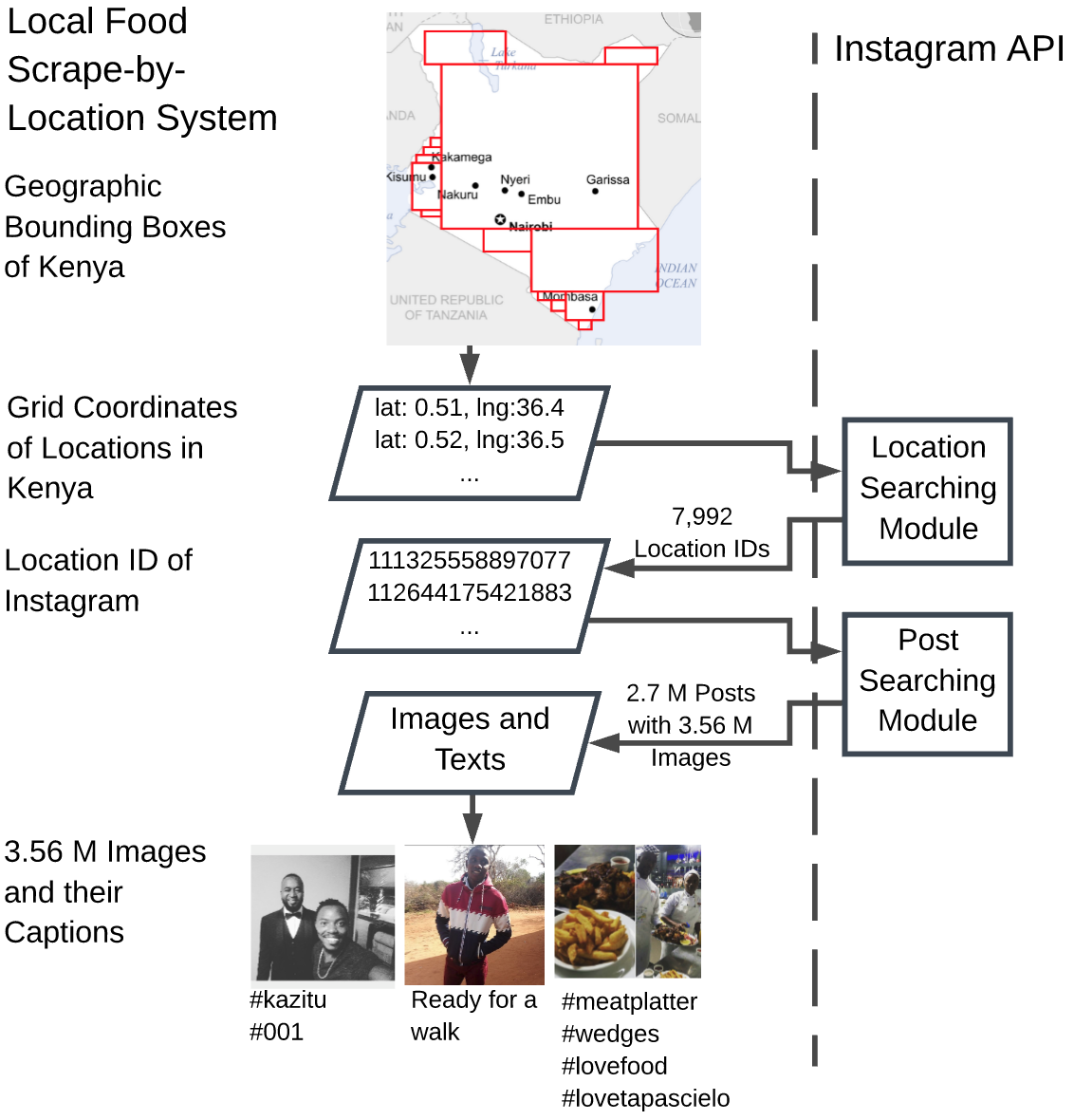}
\end{minipage}
\caption{Overview of the proposed Scrape-by-Keywords System (left) and Scrape-by-Location System (right)}
\label{fig:search}

\end{figure}

In this section, we first describe two multimodal dataset collection methods, then the resulting datasets, and finally the classifiers we designed to analyze them.
We developed the ``scrape-by-location'' method to collect images on Instagram from Kenya and the ``scrape-by-keywords'' method to collect popular Kenyan food images posted on Instagram, but not necessarily from Kenya.  Our systems are shown in Figures~\ref{fig:search}.




\subsection{Scrape-by-Keywords}

Our scrape-by-keywords data collection method relied on Kenyan experts to provide us with a list of foods that are popular in Kenya.  
We received a list of 38 food names in the Kiswahili language.
We used the keyword searching module of the Instagram Scraping API~\cite{instagramapi} to search for 
Instagram posts that included at least one of the 38 food names in their image captions (usually tagged with a hashtag) and downloaded both images and captions.  The API provided us with $\sim$40,000 data points and stopped finding additional posts after one day of search (March 23--24, 2019).  
We manually inspected and filtered out about 10,000 images that did not include any food.  Our scrape-by-keywords process thus resulted in about 30,000 Kenyan food images and their captions, which include the names of the foods in Kiswahili.

\setcounter{page}{3}

\subsection{Scrape-by-Location} 

We defined a set of rectangular regions on the map of Kenya (red boxes in Fig.~\ref{fig:search}) to cover its territory and then defined a grid within these regions.  The grid has a stride of 0.02 degree of longitude and 0.02 degree of latitude.   For each point on the grid we searched all nearby locations registered on Instagram using the location searching module of the Instagram Scraping API.  We recorded the identifier (ID) of each retrieved location.  Using the ID, we then applied the Instagram post searching module of the API to retrieve recent posts that had been uploaded from the location with that ID.  For each post, we downloaded as much information as available.  This included the primary key, ID, image(s), URL(s), and potentially caption, and latitude and longitude of the location.  We applied our  scrape-by-location system for 20 days, spanning March 7 to March 27, 2019, and retrieved 2.7 million Instagram posts. The posts contained a total of 3.56 million images, many with captions.

\setcounter{page}{4}

\subsection{Kenyan Food Type Dataset {\em KenyanFood13}}

To develop a food type dataset, we analyzed $\sim$30,000 images that we collected with our scrape-by-keywords system.  We noticed that the number of images per food type differed significantly.  We collected 9,142 images of bhaji, but only 316 images of mukimo. 
Only 15 of the 38 food types were represented by more than 500 images. 
Moreover, there is no guarantee that the images downloaded by the scrape-by-keywords system actually include the types of food that their captions mention. All we know is that the captions include at least one of the 40 Kenyan food names that we used as keywords in the collection process. Considering these issues, 
we decided to reduce the dataset to only 13 classes, where each class has at least 200 samples.  We manually inspected each remaining image to ensure that it is a photograph of the food type that it was assigned to. This process resulted in 8,174 images in 13 food type classes.  Sample images of our KenyanFood13 dataset and the number of images in each of the 13 food type classes are shown in Fig.~\ref{fig:type_example}.

\subsection{{\em Kenya104K} Dataset}

In order to discover eating patterns based on analyzing large datasets (like the 3.5 million multimodal dataset we collected with our scrape-by-location method),
health informaticists need a tool to distinguish images of food from non-food. To create such a tool, a classifier needs to be trained on images that are categorized into food and non-food. To create a dataset automatically (versus by crowdsourcing), a classifier is needed. To resolve this causality dilemma, we build up our training set in piecemeal. We first assign the 30,000 images retrieved by our scrape-by-keywords system to the class of food images.
We also use 30,000 non-food images from our scrape-by-location collection, which we selected manually. 
We then combine them with 9,658 images from the Instagram Food Dataset (IFD) and 53,572 images from FCD (food and non-food). 
Then, we trained a food/non-food detector that detects the food images collected from Kenya. By applying the trained food detector, we initially detected about 70,000 food images and after manual inspection, we ended up with 52,000 food images. To create negative samples, we added 52,000 manually inspected random samples from the images collected by ``scraping-by-location.'' 

To develop our classifier {\bf KenyanFC}, which can distinguish between food and non-food content in images, we
fine-tuned ResNeXt101~\cite{XieGiDoTuHe17}, pre-trained on ImageNet dataset~\cite{DengDoSoLiLiLi09},
with a merged dataset containing FCD~\cite{KagayaAi15}, Food-5K~\cite{SinglaYuEb16}, as well as our own food/non-food dataset KenyanFood104k.
 The number of nodes of the output layer of ResNeXt101 was changed to two to adapt to the food/non-food classification task.
\begin{figure}[h]
    \centering
    \includegraphics[width=0.8\columnwidth]{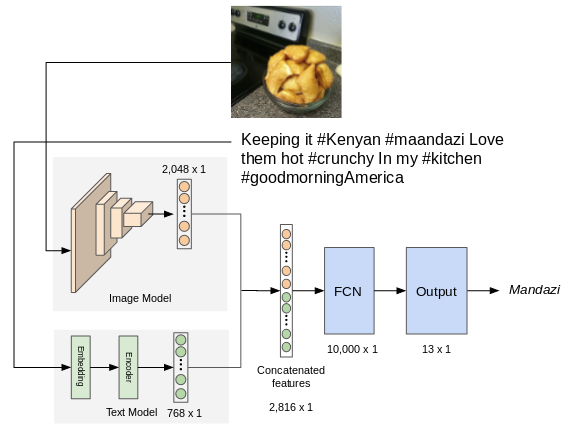}
    \caption{Architecture of food type recognition model (FCN stands for fully connected network).}
    \label{fig:mesh2}
\end{figure}

\subsection{Our Classifier {\em KenyanFTR}}

To develop a classifier that can distinguish among 13 Kenyan food types, 
we propose a feature fusion model that extracts features from each image and its corresponding caption. The system architecture of our model KenyanFTR is shown in Figure~\ref{fig:mesh2}.
\subsection{Our Classifier {\em KenyanFC}}

The model uses the BERT language model~\cite{DevlinChLeTo19} to extract features from the Instagram image captions and fuses these features with the extracted image features. We passed each input token --the words in the caption-- through three embedding layers (token embedding, segment embedding, and position embedding) and these three embedding representations are summed element-wise to produce a single input representation. The input representation is passed to the encoder layer of BERT, and we used the outputs as features representing text.
To represent the image, we applied the ResNeXt101 model pre-trained on ImageNet and extracted a feature vector from the last hidden layer. After we extracted features from both BERT and ResNeXt101, we concatenated the two feature vectors into a new vector followed by a hidden layer with 10,000 neurons and an output layer with 13 classes.

\section{Experimental Results}
 We first introduce experiments with our classifier KenyanFC on our food/non-food dataset Kenya104K followed by experiments with our multimodal model KenyanFTR on the KenyanFood13 dataset. Finally, we apply our KenyanFTR model and other methods to thousands of Instagram posts to analyze food trends in Kenya.

\subsection{Experiments for Food/Non-food Detection}

We first fine-tuned ResNeXt101 (pre-trained on ImageNet) with either Food-5K, FCD, or Kenya104K. 
For all datasets, 90\% of the images were used for training and validation (72\% train, 18\% validation) in five folds (five different splits).  
The resulting five models were tested on a hold-out test set of 10\% of the dataset.  Accuracy mean and standard deviations were then reported as averages of these five models (Table~\ref{table:fnf_result}).
\begin{center}
\begin{table}[h]
\centering
\caption{
Accuracy of food/non-food classification.}
\begin{tabular}{ |c|c|c|c| } 
 \hline
Dataset & Food-5K & FCD     & Kenya104K   \\ \hline
Food-5K & 99.24\%$\pm$ 0.08\% & 98.54\% $\pm$ 0.12\% & 95.20\% $\pm$ 0.91\% \\ \hline
FCD     & 98.44\%$\pm$ 0.45\% & 99.52\% $\pm$ 0.02\% & 95.90\% $\pm$ 0.39\%\\ \hline
Kenya104K   & 98.32\%$\pm$ 0.20\%  & 97.89\%$\pm$ 0.37\% & 99.03\%$\pm$ 0.06\% \\ \hline
Combined   & N/A  & N/A & 99.01\%$\pm$ 0.06\% \\ \hline
\end{tabular}
\label{table:fnf_result}
\caption*{Table entry $(i,j)$ means that the classifier was trained on dataset $i$ and tested on dataset $j$.}
\end{table}
\end{center}

To train our KenyanFC, we built a training set by merging the training, validation and testing set of Food-5K and FCD with the training and validation set of Kenya104K. Finally KenyanFC was evaluated on the testing set of Kenya104K.  During the training phase, we applied data augmentation including random rotation, horizontal flipping, and color jitter. During the fine-tuning process, we used stochastic gradient descent (SGD) as our optimizer with a learning rate of 0.0001 and a momentum of 0.9. To avoid over-fitting, we chose the model producing the highest validation accuracy as the final model within 10 epochs of the fine-tuning process.  Our results show high accuracy in food detection (above 95\% in all cases shown in Table~\ref{table:fnf_result}). 
The testing accuracy of our KenyanFC (99\%) shows that it generalized better when it evaluated images in the Kenyan context. 


\subsection{Experiments for Food Type Recognition}

\begin{figure}[t]
    \centering
    \includegraphics[width=0.45\columnwidth]{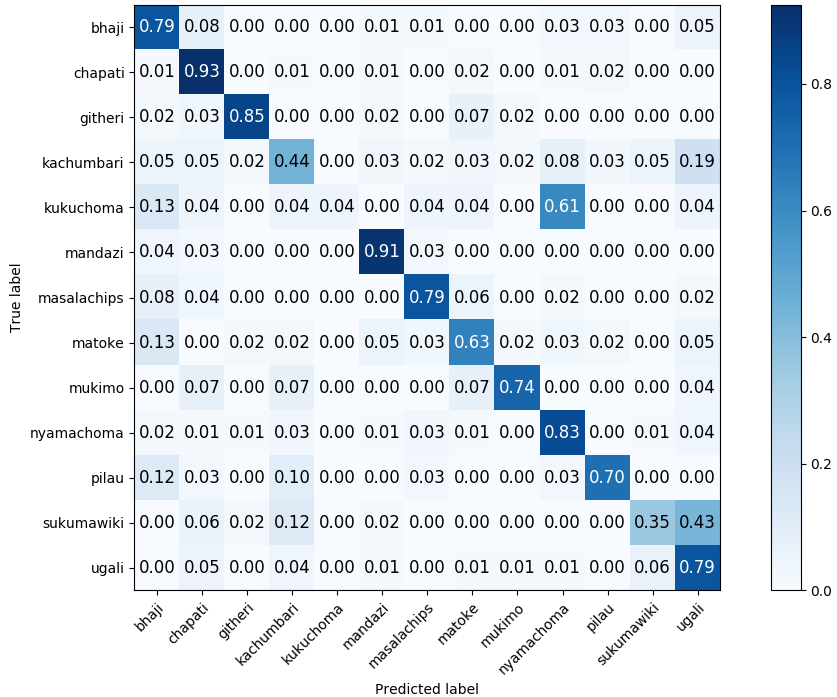}
    \includegraphics[width=0.45\columnwidth]{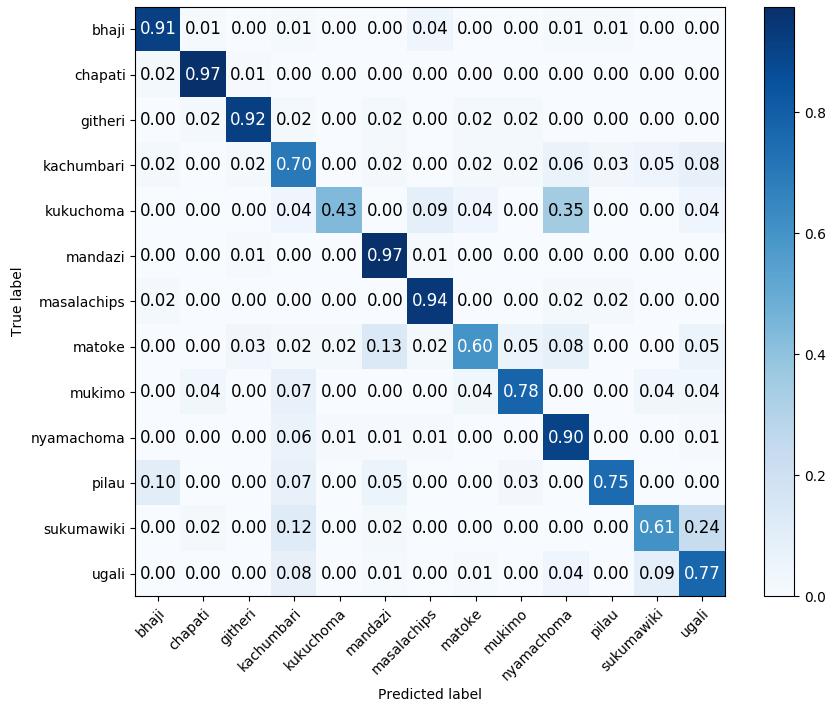}
    \caption{ KenyanFTR applied to KenyanFood13: Confusion matrices of food type recognition model based on only images (left) and based on images and captions (right). }
    \label{fig:cm_no_text}
\end{figure}

Our classifier KenyanFTR yields a top-1 accuracy of almost 81\% when tested in a 5-fold cross-validation manner on KenyanFood13.  For each class, most predicted labels match the ground truth labels, as can be seen in the confusion matrix in Fig.~\ref{fig:cm_no_text}.  For six classes, recognition accuracy surpassed 90\%.  
For comparison, we also present the confusion matrix for the classifier that only interprets images.

We note that, by design, the scrape-by-keywords data-collection method only harvests images with captions that contain the names of the Kenyan foods we targeted in our scraping.  For such images, to identify the food type, a classifier would not even have to analyze the image and could just evaluate its caption.  We found that 
a pre-trained BERT model (we tested the uncased version) reaches an accuracy level of about 98\% in identifying the food type from the captions alone. This means the model can almost always make correct predictions when food names are available.
It would be wrong, however, to assume that any Kenyan who posts food images on Instagram also includes food names in their captions. In fact, we found that for only 1,914 images out of the 52,000 images that we obtained with our scrape-by-location method, the captions contain food names.   
To train our KenyanFTR classifier to identify a food type in an image without relying on the food type name appearing in the caption, we removed any hashtagged food names from the captions in our images before using them in our experiments.

\begin{table}[h]
\centering
\caption{\label{tab:compareResult1}
Results of Ablation Studies: Accuracy of different input settings on KenyanFood13.}
\begin{tabular}{|c|r|r|}
\hline
\multirow{2}{*}{\textbf{Method}}          & \multicolumn{2}{c|}{\textbf{Test Accuracy}}                \\ \cline{2-3} 
                                 &\textbf{Top-1} & \textbf{Top-3} \\ \hline\hline
Image only          &    73.18\%$\pm$ 0.79\%  &  92.04\%$\pm$ 0.44\% \\ \hline
Caption only                &    65.30\%$\pm$ 1.70\%     &   83.68\%$\pm$ 1.55\%   \\ \hline\hline
{\bf Ours:} Image + Caption               &    81.04\%$\pm$ 0.86\%     &  95.95\%$\pm$ 0.44\%   \\ \hline
\end{tabular}
\end{table}

In order to explore the value of taking advantage of the two modalities, image and text, in our KenyanFood13 dataset, 
we conducted ablation studies with models that take as input only images or only text (Table~\ref{tab:compareResult1}). For the former, we fine-tuned a ResNeXt101 (pre-trained on ImageNet) only with the images of KenyanFood13 and evaluated its performance, while for the latter, we fine-tuned a pre-trained BERT-based model using only the captions of KenyanFood13. Finally, we compared their performance with our KenyanFTR model, which takes both images and text as input. The gain of accuracy in using both image and text modalities is significant -- the additional use of text improves the top-1 accuracy result by more than 9 percent points, while the additional use of the image improves the top-1 accuracy by 30 percent points.

We also investigated the performance of different image feature extractors.  
We compared the ResNeXt101 feature extractor used by our KenyanFTR with other popular pre-trained deep learning models, including ResNet101~\cite{HeZhReSu16}, InceptionV3~\cite{SzegedyVaIoShWo16}, InceptionV4~\cite{SzegedyIoVaAl17}, and DenseNet161~\cite{HuangLiMaWe17}. 
We note that the feature vectors computed by 
ResNet101, InceptionV3, InceptionV4, and DenseNet161 have lengths 2,048, 2,048, 1,536, and 2,208, respectively.

During the fine-tuning process in the experiment, we applied 5-fold cross-validation to train and test  our KenyanFood13 dataset. Further, we applied the same data augmentations as in our food/no-food classifier (random rotation, flipping, color jitter). 
During the fine-tuning process, we used SGD as the optimizer with a learning rate of 0.0001,  momentum of 0.9, and 12 epochs.
The top-1 testing accuracy of the different models is below 80\% except for our KenyanFTR, which achieves an accuracy of 81\% (Table~\ref{tab:compareResult2}).

\begin{table}[h]
\centering
\caption{\label{tab:compareResult2}
Results of Comparison Experiments: Accuracy of different models on KenyanFood13.}
\begin{tabular}{|c|r|r|}
\hline
\multirow{2}{*}{\textbf{Method}}          & \multicolumn{2}{c|}{\textbf{Test Accuracy}}                \\ \cline{2-3} 
                                 &\textbf{Top-1} & \textbf{Top-3} \\ \hline\hline

InceptionV3+BERT                          & 71.92\%$\pm$ 1.52\%                 & 88.57\%$\pm$ 0.68\%                 \\ \hline
InceptionV4+BERT                               & 67.40\%$\pm$ 1.49\%                 & 85.05\%$\pm$ 1.93\%                 \\ \hline
ResNet101+BERT                                 & 76.74\%$\pm$ 2.02\%                 & 93.71\%$\pm$ 1.18\%               \\ \hline
DenseNet161+BERT                         & 79.02\%$\pm$ 0.96\%                 & 95.14\%$\pm$ 0.73\%                 \\ \hline
\hline
{\bf Ours:} ResNeXt101+BERT
               &    81.04\%$\pm$ 0.86\%     &  95.95\%$\pm$ 0.44\%    \\ \hline
\end{tabular}
\end{table}

\subsection{Analysis of Food Trends in Kenya}
To investigate food preferences in Kenya (i.e., Instagram sharing preferences), we designed a tool to recognize food types in the 52,000 food images of the Kenya104K dataset. Considering the fact that we conducted an intensive grid search within the geographic boundaries of Kenya, which ensured broad coverage during the collecting process, we suggest that our dataset is sufficiently representative for research of food trends on Instagram in Kenya.

To identify food type images uploaded in Kenya, we first applied our KenyanFTR model on the 52,000 food images of the Kenya104K dataset. Because KenyanFTR was trained on KenyanFood13, which includes data of 13 popular Kenyan foods, we reason that a confidence score of at least 70\% of a food type label predicted by KenyanFTR on these images is likely correct. By visual inspection of the food images of Kenya104K, we also noticed that fruits and some Western foods such as cake and pizza are popular in Kenyan Instagram uploads.  We therefore applied YOLO v3~\cite{RedmonFa18}, pre-trained on MSCOCO~\cite{LinMaBeHaPeRaDoZi14}, to detect fruits and Western foods in the food images of Kenya104K.  We report that 25,865 images were predicted to depict food. This includes 13,975 images with Kenyan food items, 2,530 images with fruits and vegetables, and 13,860 with Western foods (details in Table~\ref{table:detected_list}). Note that for a single image, the classifiers may yield multiple prediction results since the images may indeed contain more than one type of food. 

\begin{center}
\begin{table}[h]
\centering

\caption{Some of the food types detected in Kenyan Instagram posts collected during 20 days in March 2019.}
\begin{tabular}{lr|lr } 
 \hline
Food type & \# images
&
Food type & \# images
\\ \hline
cake & 7,559 &
nyamachoma & 3,220\\
kachumbari & 2,990 &
mandazi & 2,671 \\
pizza & 2,456 &
sandwich & 2,407\\
masalachips & 1,315 &
githeri & 894\\
doughnut & 851 &
pilau & 728\\
carrot & 702 &
banana & 628\\ \hline
\label{table:detected_list}
\end{tabular}
\end{table}
\end{center}

\section{Discussion of Results}
\begin{figure}[h]
    \centering
    \includegraphics[width=0.8\columnwidth]{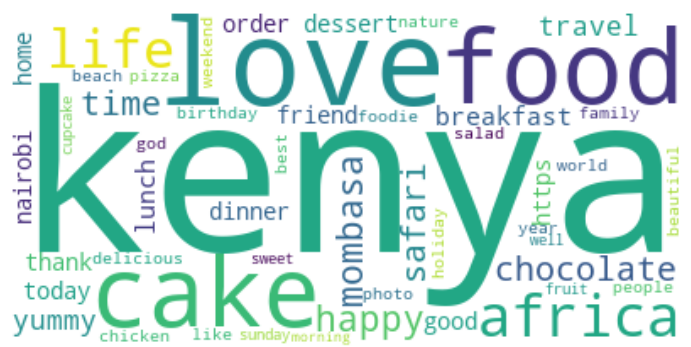}
    \caption{The word cloud shows words in captions
 of the 52K food images in Kenya104K, resized proportionally to how frequently they occurred.
    }
    \label{fig:wordcloud}
\end{figure}

Torralba et al.~\cite{TorralbaEf11} discussed the concept of dataset bias -- it is inevitable that datasets have intrinsic features that may be difficult to recognize by humans, for example, due to selection, image capture, or category-preference biases.  We suggest that this could be studied as ``a feature not a bug'' for our datasets.  
The high occurrence of the word ``love'' in image captions of Kenya104K , for example, seems to indicate that the Instagram users prefer to upload images of foods that they love (for additional examples, see Fig.~\ref{fig:wordcloud}).
A capture bias is most likely present in our data since people typically center objects of interest in a photograph.  The range of numbers of images per each food type that we collect indicates a category-preference bias.  Other challenges are the existence of multiple food items in the images, image quality (low resolution or out-of-focus), as well as unknown food types.

\subsection{Discussion of Food/Non-food Detection}

Supervised learning systems typically perform better when training and testing images belong to subsets of the same dataset.  This is also true for our classifier {\em KenyanFC}, as can be seen in Table~\ref{table:fnf_result}. From this table, we observe that, when tested on Kenya104K, models trained on Food-5K and FCD performed somewhat worse than a model trained on only Kenya104K, or a model trained on the combination of all three datasets. Also, the difference between accuracy of models trained on Kenya104K and the combination of datasets is not statistically significant (i.e., the difference in mean accuracy is smaller than the standard deviations in accuracy).    

\begin{figure}[b]
\centering
\captionsetup[subfigure]{labelformat=empty}
   \subfloat[ \label{fig:PKR}]{%
      \includegraphics[width=3.2cm, height=3.2cm]{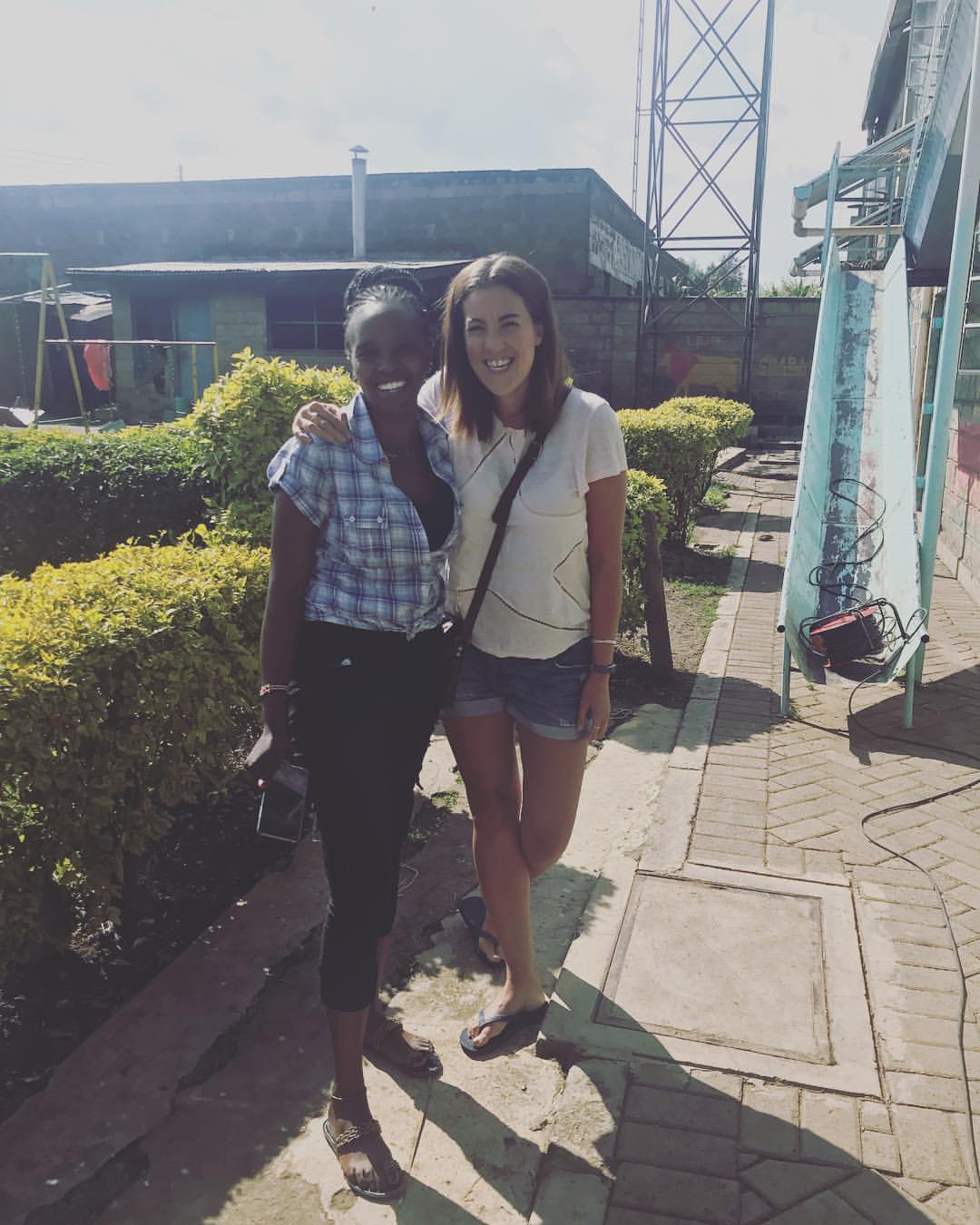}}
\hspace{1em}
   \subfloat[\label{fig:PKT} ]{%
      \includegraphics[width=3.2cm, height=3.2cm]{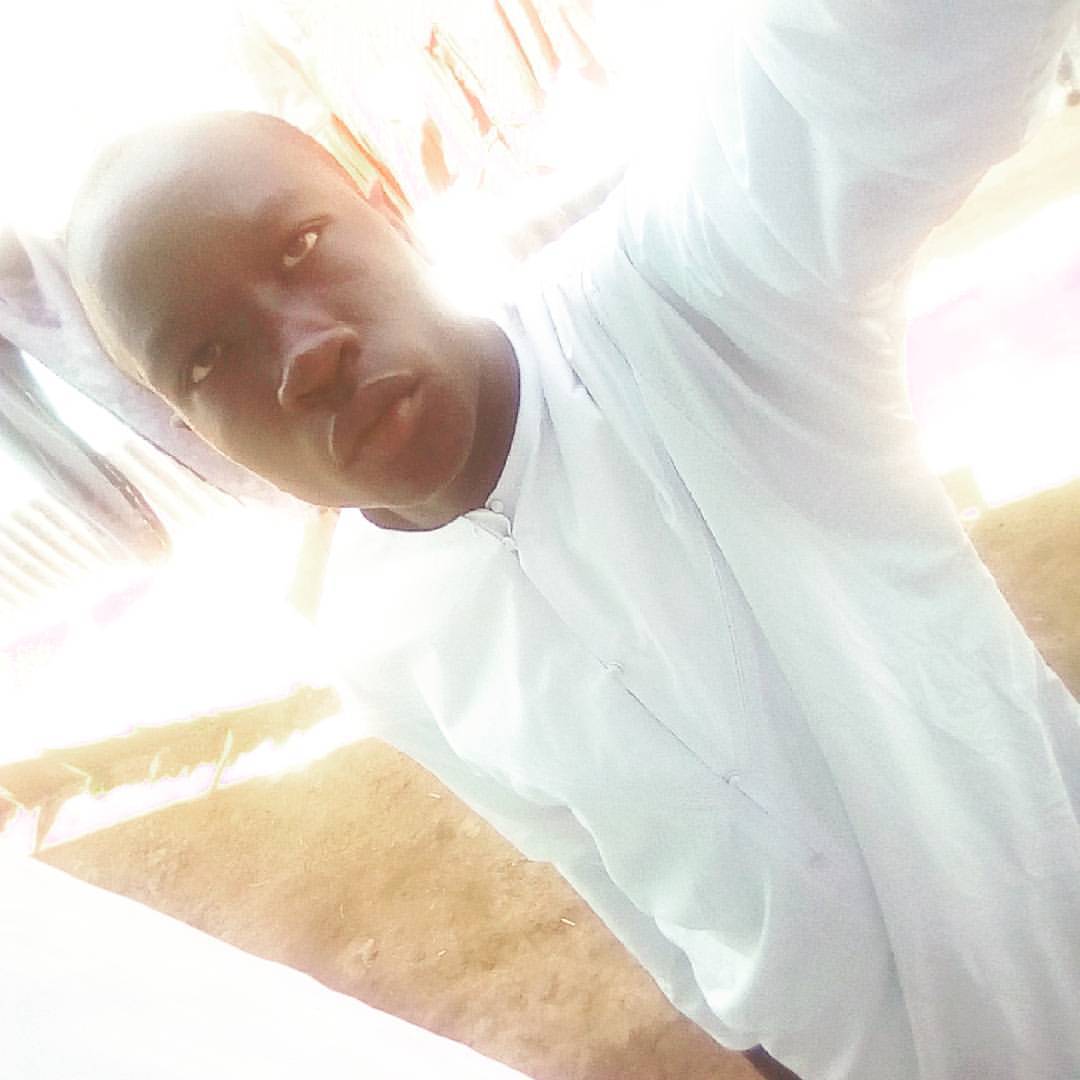}}
\hspace{1em}
   \subfloat[\label{fig:tie5}]{%
      \includegraphics[width=3.2cm, height=3.2cm]{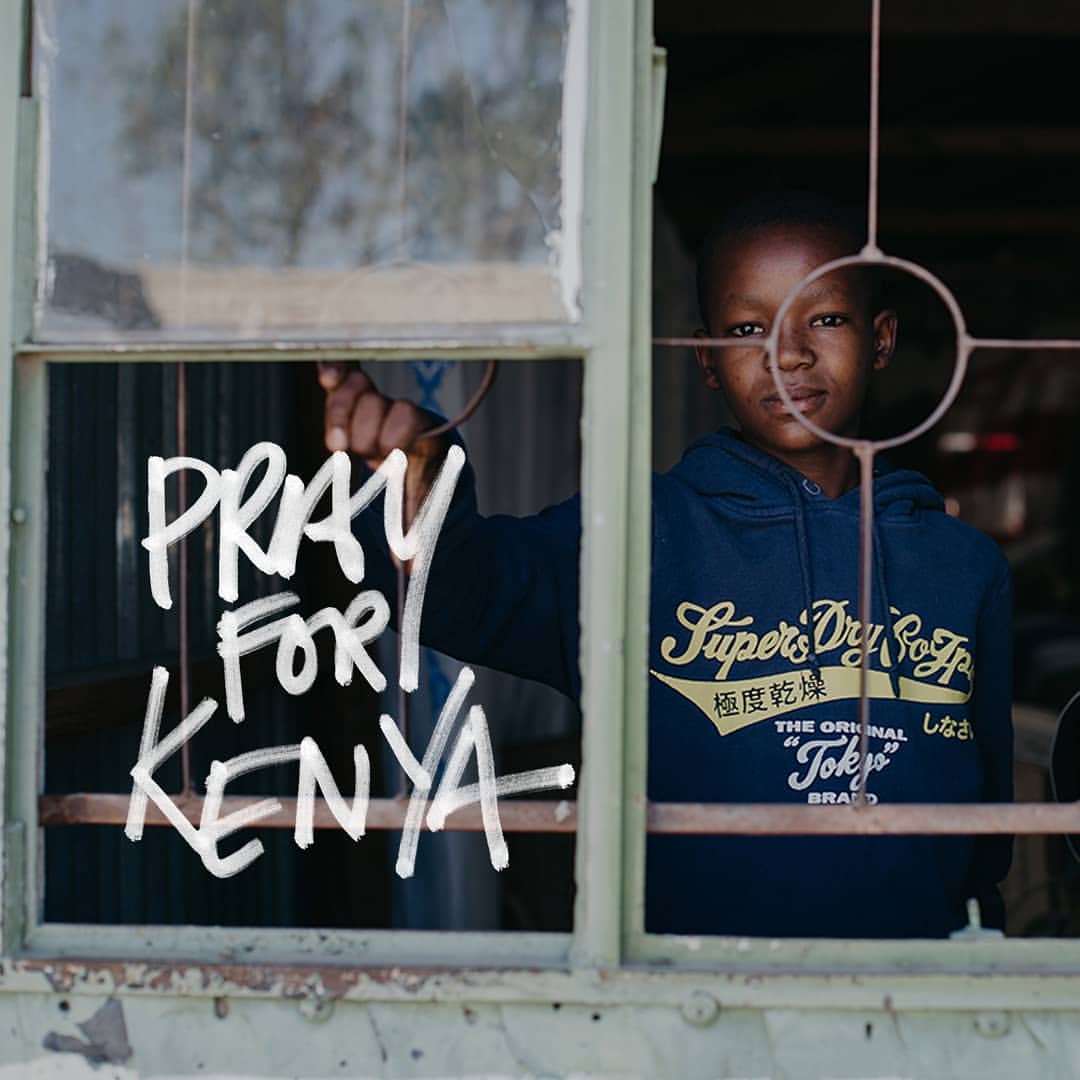}}
\hspace{1em} 
    \subfloat[\label{fig:PKR}]{%
      \includegraphics[width=3.2cm, height=3.2cm]{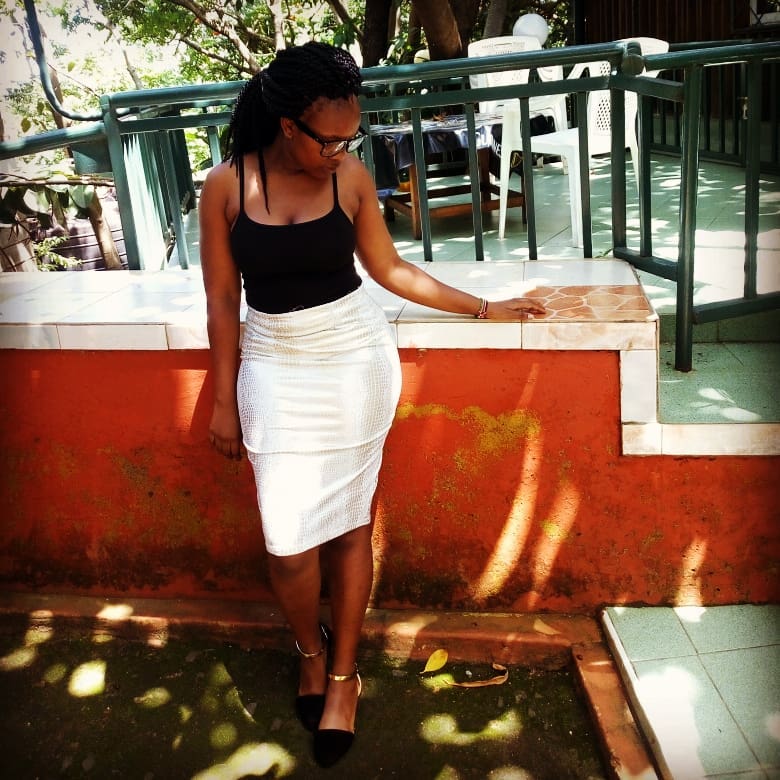}}\\
[-3ex]
      \subfloat[ \label{fig:PKR}]{%
      \includegraphics[width=3.2cm, height=3.2cm]{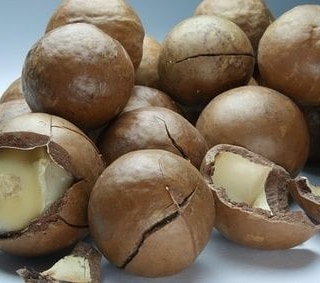}}
\hspace{1em}
   \subfloat[ \label{fig:PKT} ]{%
      \includegraphics[width=3.2cm, height=3.2cm]{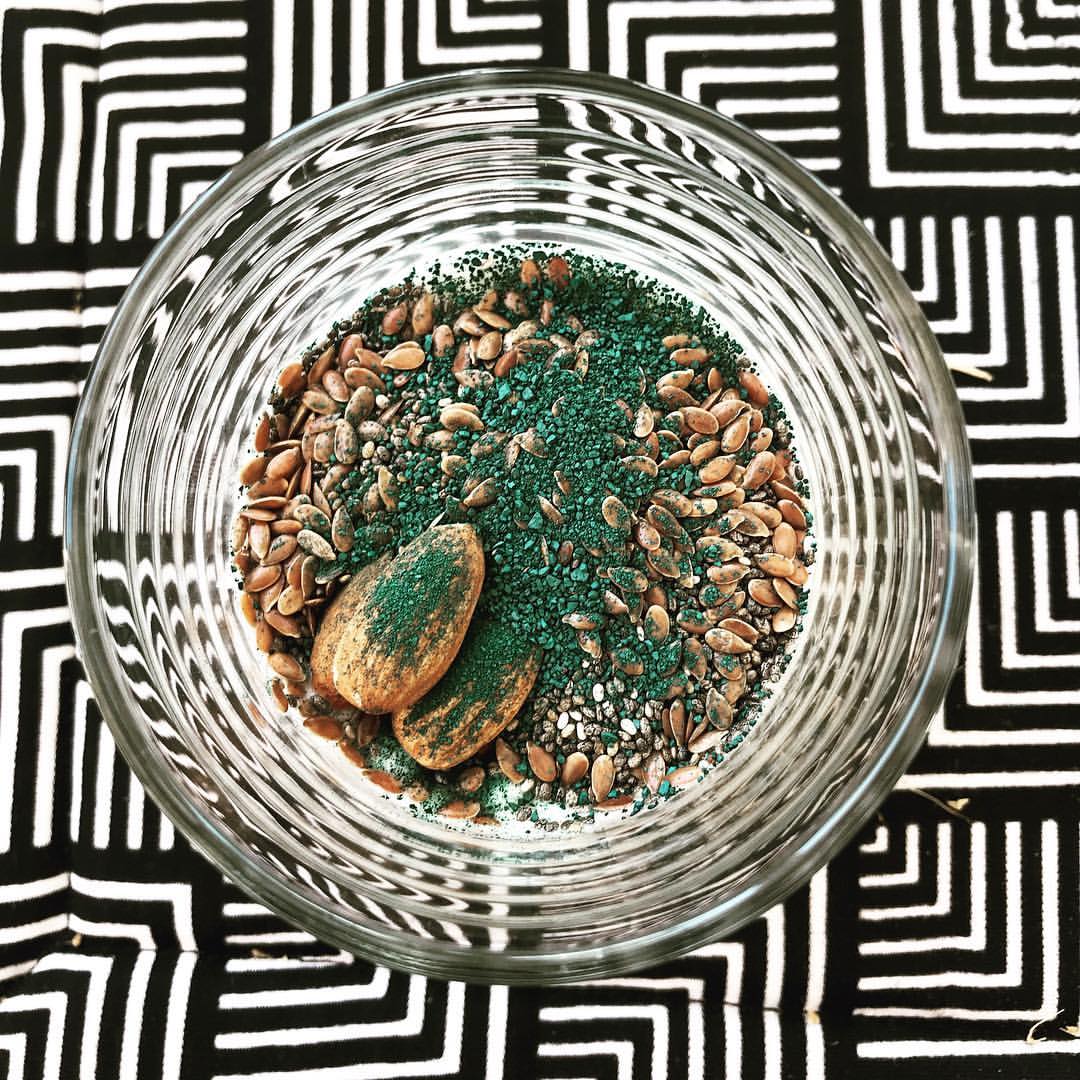}}
\hspace{1em}
   \subfloat[ \label{fig:tie5}]{%
      \includegraphics[width=3.2cm, height=3.2cm]{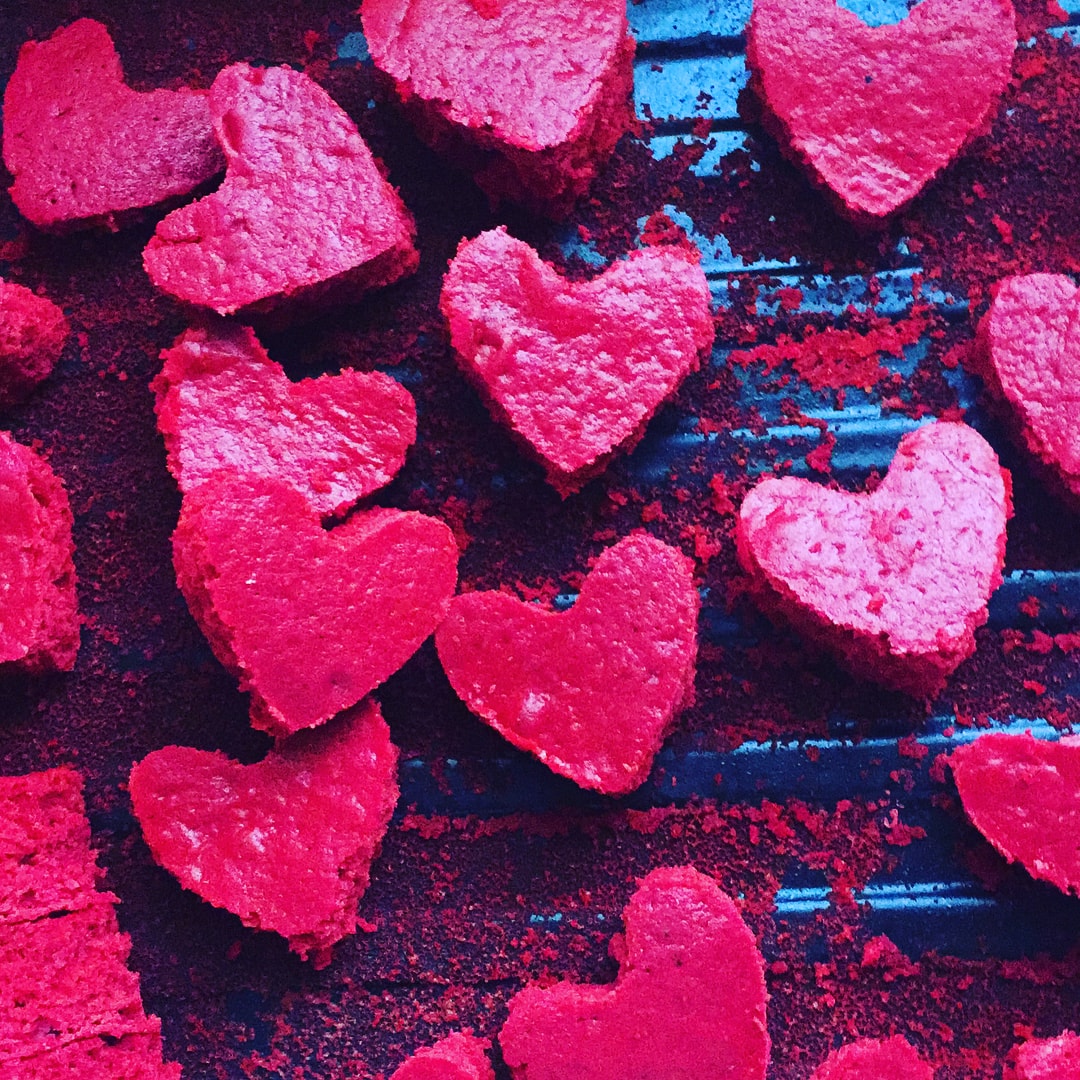}}
\hspace{1em}
    \subfloat[\label{fig:PKR}]{%
      \includegraphics[width=3.2cm, height=3.2cm]{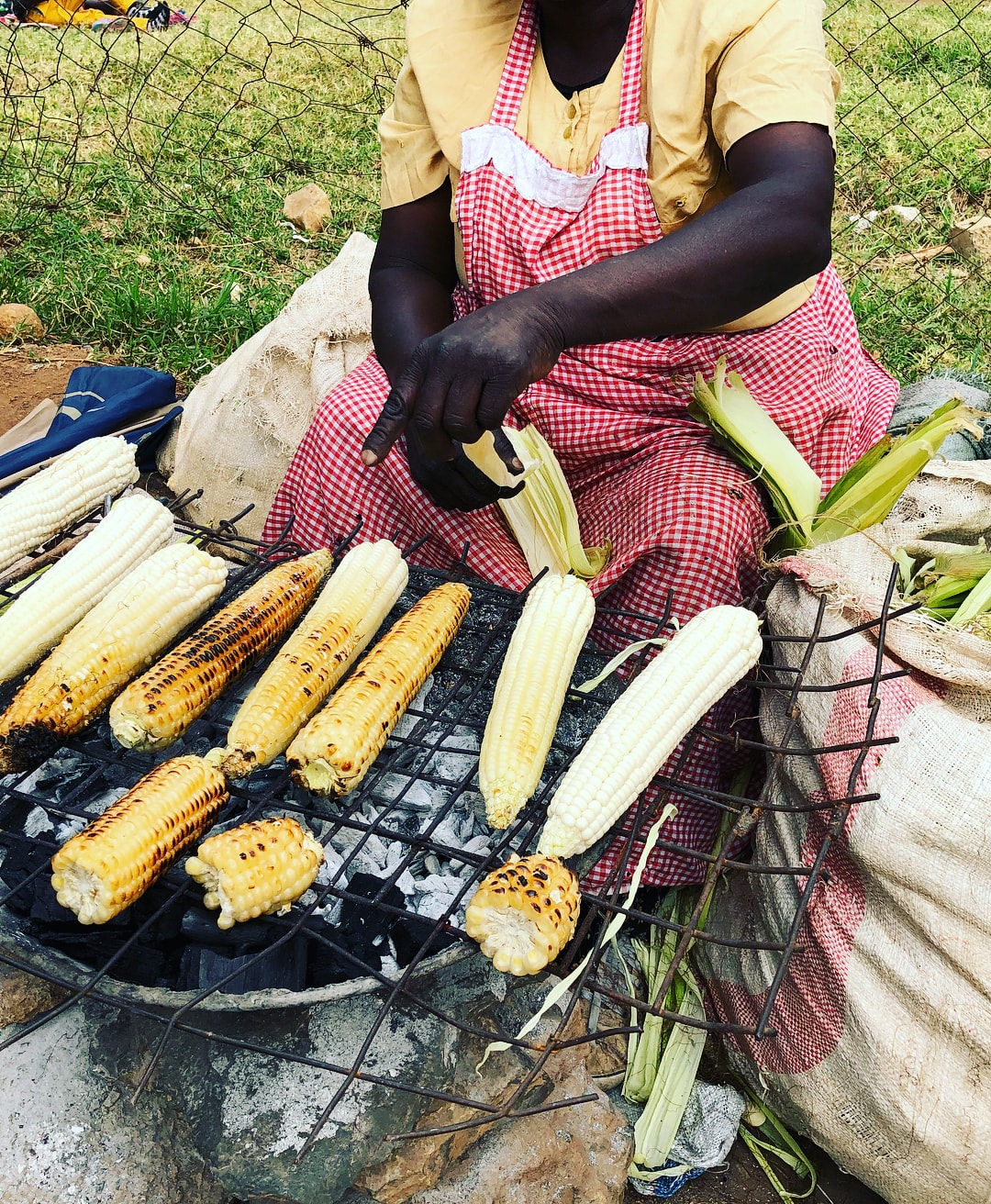}}
\caption{Images correctly classified by KenyanFC but misclassified by ResNeXt101 when trained on FCD and tested on our Kenya104K. Images in the first row are non-food images misclassified as food images; images in the second row are food images misclassified as non-food images.}
    \label{fig:fnf_failure}
\end{figure}

To explore the advantage of Kenya104K over other food/non-food datasets when processing images from Kenya, in Figure ~\ref{fig:fnf_failure}, we provide some example images that were misclassified by the model trained on FCD but correctly classified by the model trained on Kenya104K. It is obvious that images containing 
Kenyan content are confusing to the models since they are trained on other datasets that do not contain such content. This illustrates the necessity of having a dataset dedicated to a specific country or region for image recognition tasks when trying to analyze food trends in Africa.

\subsection{Discussion of Food Type Recognition}

\begin{figure}[t]
    \centering
    \includegraphics[width=\columnwidth]{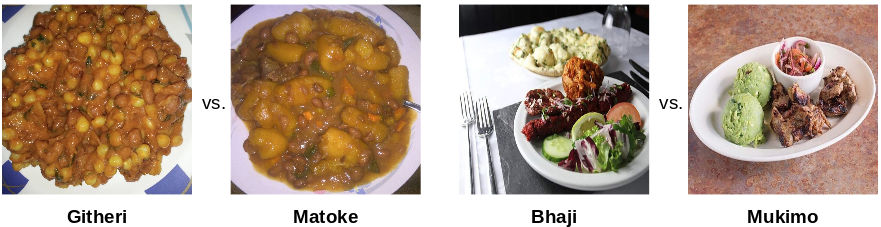}
    \caption{Examples of most confused food images in our KenyanFood13.
    }
    \label{fig:similar}
\end{figure}

Our model comparison experiments reveal that
 all five deep models fused with BERT generalized well on our dataset, with KenyanFRT performing the best  (Table~\ref{tab:compareResult2}).  For KenyanFTR, we studied the remaining challenges as follows.  For every image in a selected class, we computed their L2 distance to all other images not belonging to the current food type and found the pair of images with the smallest distance. Two examples of such similar image pairs are shown in Figure~\ref{fig:similar}. 

Our ablation study shows that the difference in accuracy of three models given images only, caption only, and images combined with captions is large (Table~\ref{tab:compareResult1}). The model trained on images achieved a top-1 accuracy of 73.18\% and top-3 accuracy of 92.04\%, while the model trained on captions achieved top-1 accuracy of 65.30\% and top-3 accuracy of 83.68\%, reflecting the fact that images are more informative when training models. However, when we combined images with their corresponding captions, recognition accuracy increased from 73\% to 81\%, which reveals the advantage of multimodality of our KenyanFood13 dataset. 
Interestingly, comparing the confusion matrices (Fig.~\ref{fig:cm_no_text}), we observe that the information brought by captions helped significantly in reducing the occurrence of misclassification.
For example, ``kachumbari'' and ``sukuma wiki'' are very likely to be recognized as ``ugali'' since they are always served and eaten with ``ugali,'' and so they frequently appear together in the images.  After including the captions into the training process, the misclassification rate of ``kachumbari'' and ``sukuma wiki'' with respect to ``ugali'' decreased from 19\% and 43\% to 8\% and 24\%, respectively. Also, the misclassification rate between ``kuku choma'' and ``nyama choma'' was significant simply because they look similar, but after including the captions along with the images, the misclassification rate dropped significantly from 61\% to 35\%. 

\subsection{Discussion of Food Trend Analysis}

A healthy diet is important for good health and can protect against diseases such as diabetes and cardiovascular disease~\cite{who}. According to the World Health Organization, healthy diets for adults should be high in vegetables and fruits, have less than 30\% of total energy consumption from fat, and have less than 5~g of salt content per day. Access to a healthy diet, however, can be hindered by factors such as inability to afford healthy foods and limited access to healthy food options ~\cite{IgumborSaPuTsScPuSwDuHa2012big,YngveMaHuTs2009food}.

The relationship between socioeconomic status, diet, and obesity is not always straightforward. For example, the risk for obesity can increase with wealth, as was shown in a study conducted in the Karonga District and Lilongwe city in Malawi~\cite{PriceCrAmKaMuTaBrLaMwNk2018}. In contrast, several studies have shown positive associations between lower socioeconomic status and lack of access to healthy foods and higher obesity prevalence~\cite{BlanchardMa2007retail}. Understanding changes in attitudes and sentiments towards unhealthy foods can be useful for education and implementation of interventions to improve health in communities.

\begin{figure}[t]
    \centering
    \includegraphics[width=0.5\columnwidth]{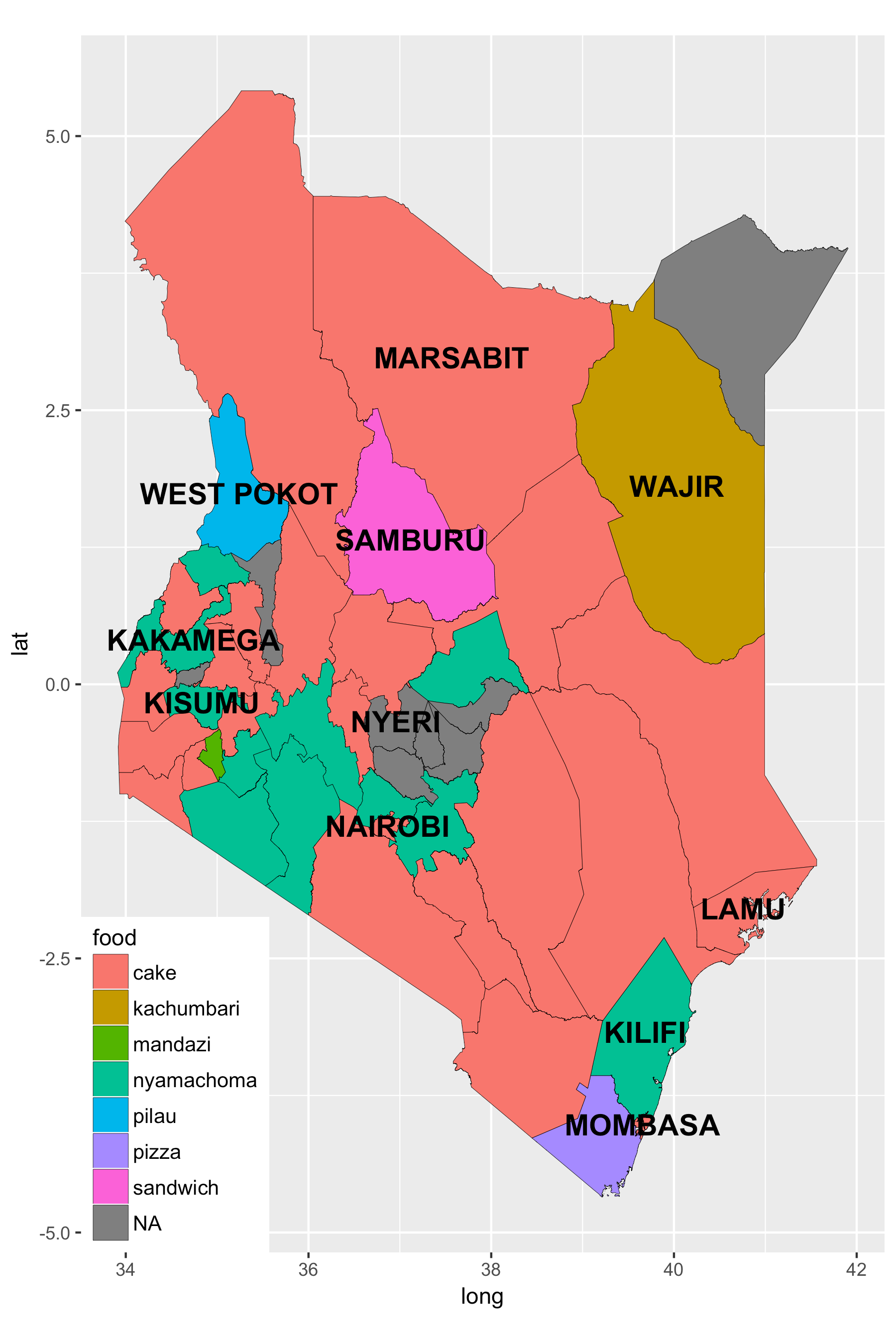}
    \caption{Most popular types of food per county in Kenya, according to 20 days of Instragram posts in March 2019.
    }
    \label{fig:favorite}
\end{figure}

 A map that reveals the foods most popular for Instagram uploads in each county in Kenya is shown in Fig.~\ref{fig:favorite}.  It is based on our analysis of the 52,000 food images of Kenya104K. Cake is the most popular food type for Instagram uploads in many counties of Kenya, as well as developed areas such as Nairobi and Mombasa. A likely reason is that people enjoy sharing images of cakes on social media, especially during celebrations (e.g., birthday cakes).
 Interestingly, other Western foods, such as pizza and sandwiches, are popular in some remote areas, while people living in south-central Kenya prefer to upload images of classic Kenyan foods such as ``nyama choma''and ``mandazi.''

To take a more intuitive look at the 52,000 food images, we created a word cloud of their captions, which is shown in Figure~\ref{fig:wordcloud}. High frequency of words like ``travel,'' ``Kenya,'' ``Africa'' may indicate that many of these posts are sent by tourists, since these words are not likely to be used by local people living in the country.

\section{Conclusions and Future Work}
In this paper, we presented two systems to scrape social media photos and their associated metadata, scrape-by-keywords and scrape-by-location, and two datasets that we developed with the help of these systems, Kenya104K for the food/non-food detection task, and KenyanFood13 for food type recognition task\st{s}. 
Extensive experiments revealed the advantages of having a Kenyan-food-specific dataset for training a classifier to detect such food and of having a multimodal dataset for classification of Kenyan food types.  We note that our food type classifier is Instagram-agnostic and can be applied to images with or without captions. We applied the classifiers, in combination with existing food-image and text classifiers, to 3.56 million images that were posted on Instagram across Kenya over a period of 20 days. As an example of the health science analysis that our work enables, we reveal that the most popular foods for Instagram uploads in Kenya were cake and roasted meat.
Social scientists may use our datasets and/or our data scraping processes and classifiers to analyze food trends, dietary values, geographical differences, the impact of tourism, etc., by   
collecting Instragram posts over additional periods of time.

Our code and datasets are publicly available and could be augmented by annotations such as cuisine type, flavors, or ingredients.  
Furthermore, a mobile phone application could be devised that would 
inform Kenyan users, in real time, of the dietary values of the meals they are eating or are interested in eating.

\noindent
{\bf Acknowledgements.} The authors thank the National Science Foundation (1838193) and the Hariri Institute for Computing and Computational Science \& Engineering at Boston University 
for partial support of this work.

\bibliographystyle{unsrtnat}  
\bibliography{references}  






\end{document}